\documentclass[letterpaper]{article}
\usepackage[preprint]{aaai2027}
\usepackage[hyphens]{url}
\usepackage{graphicx}
\usepackage{natbib}
\usepackage{caption}
\usepackage{amsmath}
\usepackage{amssymb}
\usepackage{colortbl}
\usepackage{multirow}
\usepackage{booktabs}
\usepackage{pifont}
\usepackage{microtype}
\usepackage{makecell}
\usepackage{tabularx}
\usepackage{gensymb}
\usepackage{cuted}
\usepackage{textcomp}
\usepackage{tcolorbox}
\tcbuselibrary{breakable,skins}
\usepackage{listings}
\lstdefinestyle{jsonstyle}{
  basicstyle=\ttfamily\small,
  columns=fullflexible,
  frame=none,
  xleftmargin=0pt,
  showstringspaces=false,
  breaklines=true,
  aboveskip=0pt,
  belowskip=0pt,
  escapeinside={(*@}{@*)}
}
\newcommand{\Xuser}{\textbf{X\textsubscript{User}:}}
\newcommand{\Xagent}{\textbf{X\textsubscript{Agent}:}}
\newcommand{\imgtok}{\textcolor{green!60!black}{\mbox{\ttfamily\footnotesize<image>}}}
\newcommand{\tasktitle}[1]{\noindent\textcolor{blue!70!black}{\bfseries\itshape #1}\par}
\newcommand{\frameitem}[1]{\mbox{Frame-#1:\,\imgtok}}

\begin{document}

\title{CoordRefer: Coordinate-Aware 3D Visual Grounding from Multiview Images}

\author{
Haijie Li\textsuperscript{\rm 1},
Jiaxin Zhang\textsuperscript{\rm 2},
Dave Zhenyu Chen\textsuperscript{\rm 3,*},
Youyu Chen\textsuperscript{\rm 2},
Yanmin Wu\textsuperscript{\rm 1},
Jian Zhang\textsuperscript{\rm 1,4,$\dagger$}
}
\affiliations{
\textsuperscript{\rm 1}School of Electronic and Computer Engineering, Peking University\\
\textsuperscript{\rm 2}Harbin Institute of Technology 
\textsuperscript{\rm 3}Huawei\\
\textsuperscript{\rm 4}Guangdong Provincial Key Laboratory of Ultra High Definition Immersive Media Technology,\\ Shenzhen Graduate School, Peking University\\
\textsuperscript{*}Project Leader \quad \textsuperscript{$\dagger$}Corresponding Author
\begin{center}
Project page:
{\url{https://lhj-git.github.io/CoordRefer/}}
\end{center}
\vspace{-1em}
}

\maketitle
\begin{strip}
  \centering
  \includegraphics[width=\textwidth]{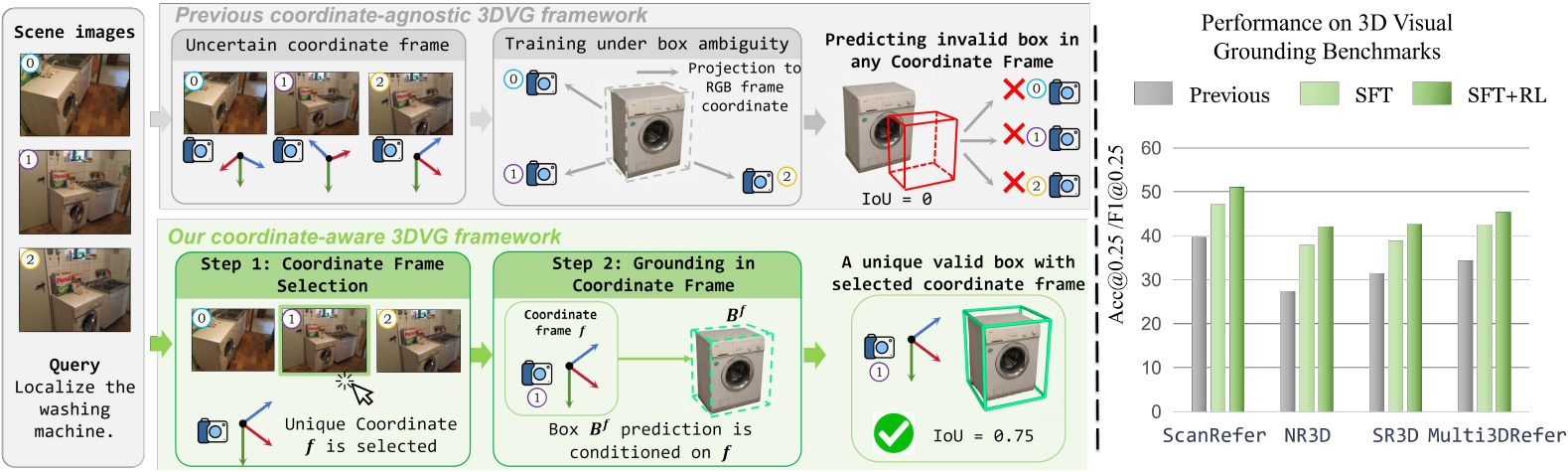}
  \captionof{figure}{
    Left: {Resolving box ambiguity for 3D visual grounding.}
    Existing coordinate-agnostic frameworks jointly infer coordinate frames and 3D boxes, causing ambiguous supervision and producing invalid coordinate--box predictions. Our coordinate-aware framework decouples these two objectives by first selecting a coordinate frame and then grounding objects in the selected coordinate system, yielding a unique box representation and improved performance.
    Right: {Performance gains on 3D visual grounding benchmarks.}
    Our framework achieves improvements of 11--14\% across multiple benchmarks over baseline.
  }
  \label{fig:teaser}
\end{strip}

\begin{abstract}
Multiview image-based 3D visual grounding predicts a coordinate frame to define a coordinate system and then regresses a 3D bounding box for localization.
However, existing methods jointly optimize coordinate frame selection and box regression, leading to coordinate-relative box ambiguity and degraded grounding performance.
This ambiguity arises because the same box admits different numerical representations across coordinate frames, creating multiple optimization targets and yielding invalid compromise predictions.
To tackle this challenge, we propose CoordRefer, a coordinate-aware framework that decouples coordinate frame selection from coordinate-conditioned grounding.
CoordRefer first selects a reference frame to define the coordinate system and then conditions 3D box prediction on the coordinate system.
We perform coordinate-aware supervised fine-tuning to establish coordinate frame selection and coordinate-conditioned box regression, followed by Group Relative Policy Optimization with 3D IoU-based rewards to align both stages with downstream grounding quality.
On ScanRefer with Qwen3-VL-2B, CoordRefer achieves gains of +11\% $\mathrm{Acc}@0.25$ and +7\% $\mathrm{Acc}@0.5$ over the coordinate-agnostic baseline, while its geometrically refined variant surpasses methods using explicit 3D inputs.
\end{abstract}

\section{Introduction}
\label{sec:intro}
The rapid advancement of multimodal large language models (MLLMs)~\cite{bai2025qwen2,bai2025qwen3,LLaVA-OneVision-1.5,hurst2024gpt,lillava} has significantly broadened their applicability across diverse domains, including complex 3D perception tasks. 
Among these, 3D visual grounding~\cite{chen2020scanrefer,achlioptas2020referit_3d}, requiring the model to understand a natural language description and accurately localize the corresponding object in 3D space, has been extensively studied using explicit 3D inputs~\cite{huang20253d,zhu2024llava,chen2024grounded3dllm,hong20233d,zheng2025video}. 
Although these methods achieve strong performance, their scalability remains under debate due to the data scarcity of 3D scenes. 

\begin{figure}[t]
  \centering
  \includegraphics[width=\columnwidth]
    {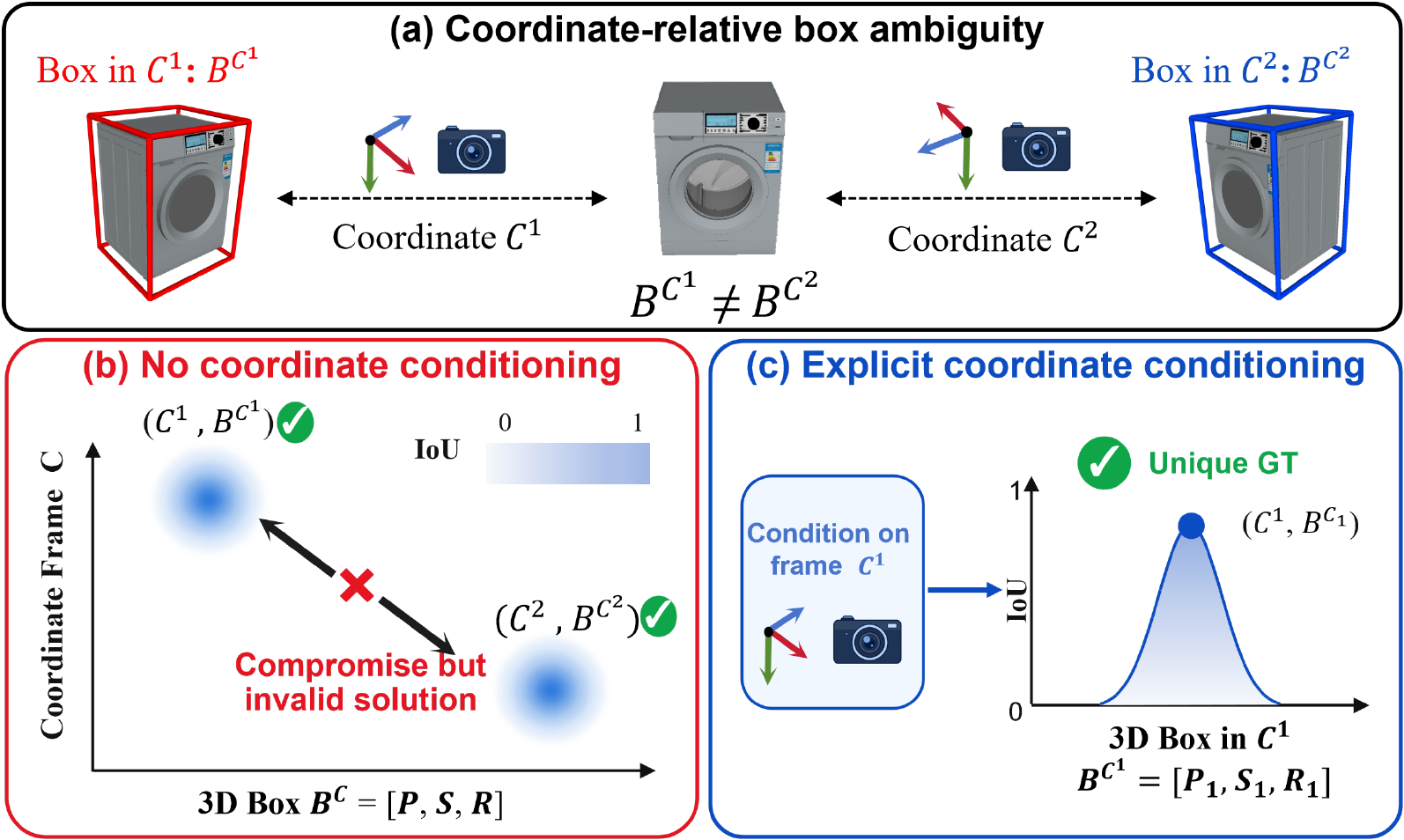}
    \vspace{-4mm}
    \caption{Coordinate-relative box ambiguity and its resolution through explicit coordinate conditioning.
(a) The same physical 3D box has different representations under different coordinate frames.
(b) Without coordinate conditioning, multiple equivalent targets may lead to a compromise but invalid solution.
(c) Conditioning box regression on the selected coordinate frame provides a unique optimization target.}
    \vspace{-4mm}
  \label{fig:ambiguity}
\end{figure}

Recent studies also explore a parallel route to tackle 3D visual grounding by directly producing 3D boxes from the input multiview images.
This line of work typically formalizes a joint framework by simultaneously identifying a frame to set up a coordinate system and regressing a 3D bounding box within that system~\cite{zhang2025from, zheng2025learning, hu2025omniview}.
Despite their effectiveness, we observe that this prevailing framework suffers from a fundamental weakness: jointly predicting the coordinate frame and the 3D box causes \textbf{coordinate-relative box ambiguity}. As illustrated in Fig.~\ref{fig:teaser}, this ambiguity induces multiple optimization targets, causing the model to predict an invalid box representation that often does not correspond to any coordinate frame.
As illustrated in Fig.~\ref{fig:ambiguity}, the same physical 3D box is represented by different numerical parameters under different coordinate frames. Jointly predicting the coordinate frame and 3D box therefore introduces multiple equivalent but inconsistent optimization targets. Optimizing over these targets may yield a compromise prediction that does not correspond to a valid box in any coordinate frame. Explicitly conditioning box regression on the selected coordinate frame resolves this ambiguity by reducing the optimization to a unique and well-defined target.

Based on this observation, we propose a coordinate-aware framework for image-based 3D visual grounding. As illustrated in Fig.~\ref{fig:teaser}, our framework decouples coordinate frame selection from 3D box regression. It first selects a reference frame to define the coordinate system and then explicitly conditions box regression on the coordinate system. By fixing the coordinate frame before localization, the model is optimized toward a unique box representation, thereby eliminating coordinate-relative box ambiguity.
To realize this formulation, we adopt a progressive training framework consisting of coordinate-aware supervised fine-tuning (SFT) and reinforcement learning (RL). During SFT, the model is trained to select an appropriate coordinate frame from the input views and predict the target 3D box conditioned on the selected coordinate frame. This establishes an explicit correspondence between the coordinate system and its box representation, enabling stable two-stage grounding.
We then apply Group Relative Policy Optimization (GRPO)~\cite{guo2025deepseek} to optimize both stages with 3D IoU-based rewards. Specifically, box regression is directly rewarded by the 3D IoU between the predicted and ground-truth boxes, while coordinate frame selection is optimized using the downstream 3D IoU achieved under the selected frame as an indirect reward. By directly aligning both stages with localization accuracy, this design reduces the optimization gaps introduced by heuristic frame supervision and token-level cross-entropy training.

As a result, our proposed framework achieves improvements of 11–14\% in $\mathrm{Acc}@0.25$ and 5-7\% in $\mathrm{Acc}@0.5$ over baseline across multiple 3D visual grounding benchmarks. The RGB-only model achieves competitive performance against methods using explicit 3D inputs, while its geometrically refined variant further surpasses them.

The main contributions of this work are as follows:
\begin{enumerate}
\item We identify \textbf{coordinate-relative box ambiguity} in existing image-based 3D visual grounding frameworks, where jointly predicting the coordinate frame and 3D box introduces multiple equivalent optimization targets and may lead to invalid coordinate--box predictions.
\item We propose a coordinate-aware framework that decouples coordinate frame selection from 3D box regression and explicitly conditions box prediction on the selected coordinate frame, thereby yielding a unique and stable optimization target during training.
\item We develop a progressive coordinate-aware training strategy that first establishes explicit coordinate--box correspondence through SFT and then aligns both grounding stages with localization quality through GRPO. Extensive experiments demonstrate consistent improvements across multiple benchmarks, while the geometrically refined variant surpasses methods using explicit 3D inputs.

\end{enumerate}

\section{Related Work}
\begin{figure*}[t]
  \centering
  \includegraphics[width=1\linewidth]
    {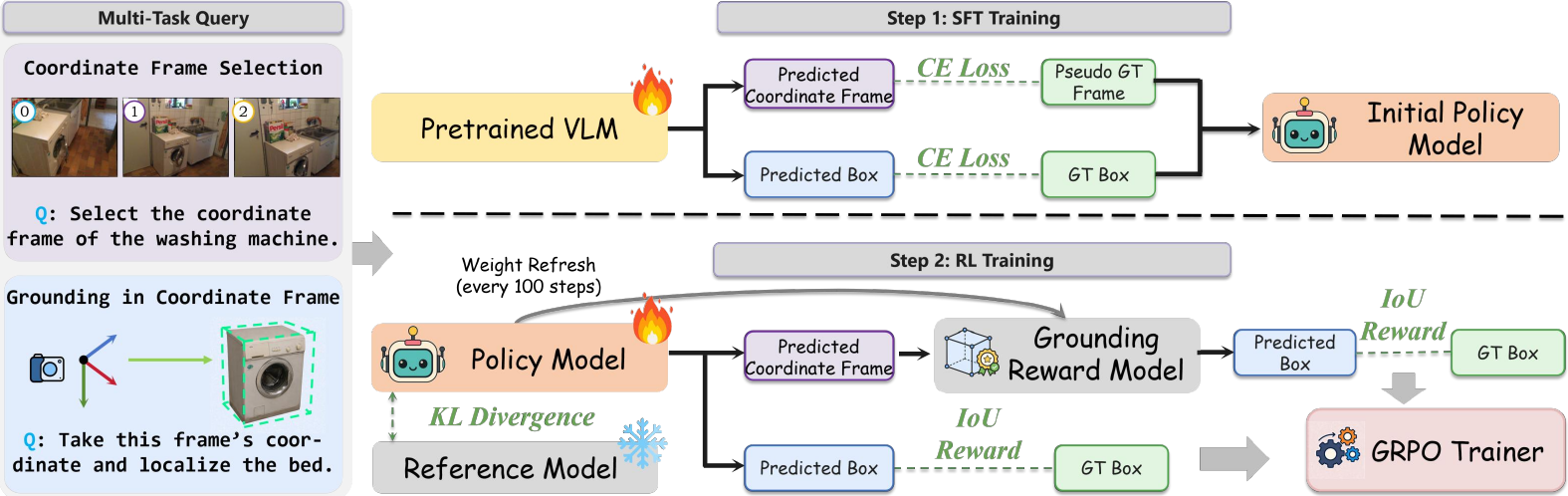}
    \vspace{-5mm}
    \caption{
    {Coordinate-aware training strategy.}
     In the SFT stage, the pretrained VLM is jointly trained on decoupled coordinate-frame selection and coordinate-conditioned grounding using pseudo frame labels and ground-truth boxes. In the RL stage, GRPO optimizes box prediction with a direct IoU reward and frame selection with an indirect IoU reward computed by a frozen grounding model. The grounding model is periodically updated with the latest policy weights.
    }
    \label{fig:main}
    \vspace{-4mm}
\end{figure*}
\subsection{Multimodal Large Language Models}

Recent multimodal large language models (MLLMs)~\cite{bai2025qwen2,bai2025qwen3,LLaVA-OneVision-1.5,hurst2024gpt,lillava,deng2025bagel} have demonstrated strong cross-modal reasoning capabilities in diverse vision-language tasks, including perception integration, language understanding, and sequential decision-making. 
Based on these foundations, recent works~\cite{shen2025vlm,li2025qinsight,xu2024fakeshield,linghu20263d,zhang2025vqinsight} further enhance MLLMs through post-training optimization~\cite{guo2025deepseek}. These approaches significantly improve visual reasoning, open-vocabulary detection, and multi-step planning performance of MLLMs. In contrast, our method focuses on using MLLMs to solve the 3D visual grounding task.

\subsection{3D Visual Grounding}

\subsubsection{Specialized 3D Grounding Models}

 Early works~\cite{chen2020scanrefer,achlioptas2020referit_3d} established benchmarks based on ScanNet~\cite{dai2017scannet} and introduced the paradigm where a 3D detector generates candidate proposals that are matched with language descriptions via cross-modal alignment. Subsequent approaches~\cite{wu2023eda} improve cross-modal correspondence through dense alignment and phrase modeling. However, these methods rely on predefined 3D proposals as an integral part of their grounding pipeline. In contrast, our core RGB-only framework directly predicts 3D bounding boxes without external proposal generators, while geometric refinement is applied only as an optional post-processing step for comparison with methods using explicit 3D information.

\subsubsection{3D Visual Grounding with MLLMs}
With the emergence of MLLMs, localization objectives can be directly integrated into the decoding process, enabling end-to-end grounding without external proposal generators. Early attempts~\cite{hong20233d,zhu2024llava,huang20253d} incorporate explicit 3D information, such as point clouds or depth maps, together with 2D visual inputs to enable multimodal reasoning for 3D grounding tasks.
Due to the cost and limited availability of point cloud data, recent efforts aim to perform 3D grounding without explicit 3D inputs. Scene-R1~\cite{yuan2025scene} introduces a pipeline that first identifies informative temporal segments and then localizes objects in image space before lifting them to 3D, though it still depends on ground-truth geometry.
SPAR~\cite{zhang2025from} predicts coordinate frames and 2.5D bounding boxes directly from video inputs, which are subsequently lifted to 3D using estimated camera parameters. Advances in geometric encoders~\cite{dust3r_cvpr24,wang2025vggt,wang2026pi} significantly improve camera pose and depth estimation from images. By incorporating such geometric priors, VG-LLM~\cite{zheng2025learning} demonstrates that models can directly predict coordinate frames and 3D bounding boxes from image sequences. 
3D-RFT~\cite{linghu20263d} further introduces reinforcement fine-tuning with verifiable frame and 3D IoU rewards, directly aligning optimization with downstream grounding performance.
These works indicate that image-based 3D visual grounding is increasingly feasible. However, existing methods largely treat coordinate selection and grounding as a coupled process. In contrast, we explicitly decouple them, enabling more stable and accurate grounding.

\section{Method}

\subsection{Preliminary}

\subsubsection{Problem Formulation}

Given a sequence of RGB images
$\mathcal{I}=\{I_t\}_{t=1}^{T}$ sampled from a video and a textual query $q$, where
$I_t\in\mathbb{R}^{h\times w\times 3}$, image-based 3D visual grounding aims to localize the referred object in 3D space without explicit 3D inputs such as depth maps or point clouds.
Following prior work~\cite{zhang2025from}, the model is required to predict a reference frame
$f\in\{1,\ldots,T\}$ and a 9-DoF bounding box
$B^{f}\in\mathbb{R}^{9}$ represented in the coordinate system of frame $f$:
\begin{equation}
(f,B^{f})=\mathcal{F}(\mathcal{I},q).
\end{equation}
Since the numerical representation of the box depends on the reference frame, a valid prediction requires the selected frame and the predicted box to be geometrically consistent.

\subsubsection{Coordinate-Agnostic Framework}

Most existing image-based 3D visual grounding approaches
~\cite{zheng2025learning,hu2025omniview,zhang2025from}
adopt a coordinate-agnostic formulation that simultaneously predicts the reference frame and the corresponding 3D bounding box:
\begin{equation}
(f,B^{f})
=
\mathcal{F}_{\mathrm{sel\_grd}}(\mathcal{I},q).
\end{equation}
However, jointly predicting the reference frame and the 3D box introduces coordinate-relative box ambiguity during optimization, often causing the model to favor a compromise yet geometrically invalid 3D box prediction.

\subsubsection{Group Relative Policy Optimization}
Group Relative Policy Optimization (GRPO)~\cite{guo2025deepseek} is a reinforcement learning algorithm that eliminates the need for a separate value critic by estimating advantages through relative comparisons among grouped responses.
Given an input $x$, the old policy $\pi_{\theta_{\mathrm{old}}}$ samples a group of $G$ responses $\{y_i\}_{i=1}^{G}$, each of which receives a task-specific reward $R_i$.
The relative advantage of the $i$-th response is computed as
\begin{equation}
\hat{A}_i =
\frac{R_i-\operatorname{mean}(\{R_j\}_{j=1}^{G})}
{\operatorname{std}(\{R_j\}_{j=1}^{G})+\epsilon},
\end{equation}
where $\epsilon$ is a small constant for numerical stability.
GRPO then optimizes the policy using a clipped likelihood-ratio objective with KL regularization against a reference policy, increasing the likelihood of responses with higher relative rewards while preventing excessive policy deviation.

\subsection{Coordinate-Aware Supervised Fine-Tuning}
\label{sec:sft}

As illustrated in Fig.~\ref{fig:main}, CoordRefer adopts a two-step training pipeline consisting of coordinate-aware supervised fine-tuning and reinforcement fine-tuning.
In the first step, we decompose frame selection and box regression into two sequential tasks with explicit coordinate conditioning:
\begin{equation}
f=\mathcal{F}_{\mathrm{sel}}(\mathcal{I},q),
\label{eq:frame_selection}
\end{equation}
\begin{equation}
B^{f}
=
\mathcal{F}_{\mathrm{grd}}(\mathcal{I}_{\mathrm{grd}},q\mid f),
\label{eq:conditioned_grounding}
\end{equation}
where $f$ denotes the selected reference frame,
$\mathcal{I}_{\mathrm{grd}}$ denotes the grounding input conditioned on $f$,
and $B^{f}$ is the 3D bounding box expressed in the coordinate system defined by $f$.
By fixing the coordinate frame before localization, each grounding sample is associated with a unique box target, thereby eliminating coordinate-relative box ambiguity.

To instantiate this formulation within a unified SFT framework, we construct two types of training instructions corresponding to Eqs.~\ref{eq:frame_selection} and~\ref{eq:conditioned_grounding}.
For coordinate-frame selection, the instruction presents the image set $\mathcal{I}$ and requires the model to predict the index $f$ of the frame used to define the coordinate system.
Its pseudo target is selected from visible candidate frames by balancing projected scale and geometric completeness.
For coordinate-conditioned grounding, the instruction takes $\mathcal{I}_{\mathrm{grd}}$ as input, designates $f$ as the coordinate reference, and requires the model to predict the 3D box in its camera coordinate system.
The supervision target is obtained by transforming the ground-truth box into the same coordinate system.
As shown in the upper part of Fig.~\ref{fig:main}, both types of instructions are jointly optimized using cross-entropy loss, yielding the initial policy for subsequent reinforcement fine-tuning.
The prompt templates and pseudo-label construction criterion are detailed in the supplementary material.
We further introduce temporal resampling and suboptimal-frame supervision to improve grounding robustness under insufficient visual observations and imperfect frame selection.

\paragraph{Temporal Resampling.}
The uniformly sampled frame set $\mathcal{I}$ used for coordinate-frame selection provides broad scene coverage but may offer insufficient observations for accurate grounding.
We therefore construct the grounding input $\mathcal{I}_{\mathrm{grd}}$ only after the reference frame $f$ is selected.
Specifically, the selected frame is placed first to define the coordinate system, followed by its temporally neighboring frames that provide complementary appearance and geometric cues.
Thus, $\mathcal{I}_{\mathrm{grd}}$ enriches the visual evidence for grounding while preserving $f$ as the unique coordinate reference.

\paragraph{Suboptimal-Frame Supervision.}
Training the grounding model only on the optimal pseudo reference frame makes it overly dependent on the quality of that frame, potentially degrading grounding performance under imperfect frame selection.
We therefore use suboptimal yet valid frames as alternative coordinate references and construct the corresponding grounding samples for additional supervision.
This exposes the model to diverse valid coordinate frames and improves its robustness to imperfect frame selection while preserving a unique box target for each sample.

\begin{table*}[t]
\centering
\setlength{\tabcolsep}{4.2pt}
\renewcommand{\arraystretch}{1.05}

\resizebox{\textwidth}{!}{
\begin{tabular}{c|cc|cc|cc|cc}
\hline

\rowcolor[HTML]{D9D9D9}
&
\multicolumn{2}{c|}{\cellcolor[HTML]{D9D9D9}ScanRefer}
&
\multicolumn{2}{c|}{\cellcolor[HTML]{D9D9D9}NR3D}
&
\multicolumn{2}{c|}{\cellcolor[HTML]{D9D9D9}SR3D}
&
\multicolumn{2}{c}{\cellcolor[HTML]{D9D9D9}Multi3DRefer}
\\

\rowcolor[HTML]{D9D9D9}
\multirow{-2}{*}{\cellcolor[HTML]{D9D9D9}Method}
&
$\mathrm{Acc}@0.25$
&
$\mathrm{Acc}@0.5$
&
$\mathrm{Acc}@0.25$
&
$\mathrm{Acc}@0.5$
&
$\mathrm{Acc}@0.25$
&
$\mathrm{Acc}@0.5$
&
$\mathrm{F1}@0.25$
&
$\mathrm{F1}@0.5$
\\ \hline

SPAR~\cite{zhang2025from}
& 31.9 & 12.4
& -- & --
& -- & --
& -- & -- \\

VG-LLM-4B~\cite{zheng2025learning}
& 36.4 & 11.8
& 25.3 & 7.5
& 28.7 & 9.8
& 31.8 & 11.2 \\

VG-LLM-8B~\cite{zheng2025learning}
& 41.6 & 14.9
& 31.5 & 11.3
& 37.2 & 14.1
& 38.6 & 14.8 \\

3D-RFT-4B~\cite{linghu20263d}
& 42.9 & 15.9
& -- & --
& -- & --
& -- & -- \\

\hline

Qwen3-VL-2B$^\dagger$ \cite{bai2025qwen3}
& 39.7 & 16.4
& 27.4 & 10.6
& 31.5 & 12.8
& 34.4 & 14.8 \\

\textbf{Ours-2B}
& \textbf{51.1} & \textbf{23.6}
& \textbf{42.1} & \textbf{17.2}
& \textbf{42.6} & \textbf{17.8}
& \textbf{45.4} & \textbf{20.8} \\

\textit{\hspace{5pt}Improvement over baseline}
&
\textit{+11.4}
&
\textit{+7.2}
&
\textit{+14.7}
&
\textit{+6.6}
&
\textit{+11.1}
&
\textit{+5.0}
&
\textit{+11.0}
&
\textit{+6.0}
\\ \hline

\end{tabular}
}
\vspace{-1.5mm}
\caption{{In-domain comparison of RGB-only methods.}
All models are trained and evaluated on the corresponding datasets.
Our method consistently outperforms existing RGB-only methods across all benchmarks. $^\dagger$ denotes the coordinate-agnostic baseline instantiated with Qwen3-VL-2B and trained under the same experimental setting.}
\label{tab:RGB_method}
\vspace{-4mm}
\end{table*}

\subsection{Coordinate-Aware Reinforcement Learning}
\label{sec:rl}

Although coordinate-aware SFT establishes frame selection and coordinate-conditioned grounding capabilities, its supervision remains misaligned with downstream localization quality.
The heuristic pseudo frame label may not identify the frame that best supports grounding, while token-level cross-entropy loss does not directly optimize box geometry.
We therefore apply GRPO to align both tasks with 3D localization performance.
As illustrated in the lower part of Fig.~\ref{fig:main}, the policy is initialized from the coordinate-aware SFT checkpoint, with a frozen copy serving as the reference model for KL regularization.
For each instruction, the policy samples a group of responses evaluated by three verifiable rewards: a shared format reward, a direct IoU reward for grounding, and an indirect IoU reward for coordinate frame selection.

\paragraph{Format Reward.}
The format reward encourages valid and parseable outputs.
A frame-selection response is valid if it contains an input frame index, while a grounding response must contain a complete 9-DoF box:
\begin{equation}
R_{\mathrm{fmt}}^{(i)}
=
\begin{cases}
1, & \text{if the output format is valid},\\
0, & \text{otherwise}.
\end{cases}
\label{eq:format_reward}
\end{equation}

\paragraph{Direct IoU Reward for Coordinate-Conditioned Grounding.}
Given a specified reference frame $f$, the policy predicts a box $\hat{B}^{f}$ in its coordinate system.
Since the corresponding ground-truth box $B^{f}$ is uniquely defined, we directly use their 3D IoU as the grounding reward:
\begin{equation}
R_{\mathrm{grd}}
=
\operatorname{IoU}_{3D}
\left(
\hat{B}^{f},
B^{f}
\right).
\label{eq:grounding_reward}
\end{equation}
Unparsable predictions receive zero reward.
Unlike token-level cross-entropy loss, this reward directly measures the 3D IoU between the predicted and ground-truth boxes.

\paragraph{Indirect IoU Reward for Coordinate Frame Selection.}
A selected frame cannot be directly evaluated by 3D IoU, while rewarding agreement with a pseudo frame label, as in 3D-RFT~\cite{linghu20263d}, retains the supervision gap of SFT.
We instead evaluate each selected frame by the grounding quality it enables.
Given a predicted frame $\hat{f}$, we construct
$\mathcal{I}_{\mathrm{grd}}^{\hat{f}}$ from $\hat{f}$ and its temporal neighbors.
A frozen coordinate-conditioned grounding model $\mathcal{F}_{\mathrm{RM}}$ predicts
\begin{equation}
\hat{B}_{\mathrm{RM}}^{\hat{f}}
=
\mathcal{F}_{\mathrm{RM}}
\left(
\mathcal{I}_{\mathrm{grd}}^{\hat{f}},
q
\mid
\hat{f}
\right),
\label{eq:reward_model_prediction}
\end{equation}
and the frame-selection reward is defined as
\begin{equation}
R_{\mathrm{sel}}
=
\operatorname{IoU}_{3D}
\left(
\hat{B}_{\mathrm{RM}}^{\hat{f}},
B^{\hat{f}}
\right),
\label{eq:frame_selection_reward}
\end{equation}
where $B^{\hat{f}}$ denotes the ground-truth box represented in the same coordinate system.
This reward measures the downstream utility of the selected frame, allowing any frame that supports accurate grounding to receive a high reward.

\paragraph{Periodically Refreshed Grounding Reward Model.}
The reward model $\mathcal{F}_{\mathrm{RM}}$ is initialized from the coordinate-aware SFT model and frozen during reward computation.
To prevent it from becoming inconsistent with the improving policy, we periodically update it using the latest policy weights:
\begin{equation}
\theta_{\mathrm{RM}}
\leftarrow
\theta,
\quad
\text{every } K \text{ optimization steps},
\label{eq:reward_model_refresh}
\end{equation}
where $K=100$ in our experiments.
This provides stable rewards within each interval while adapting frame selection to the evolving grounding capability.

\paragraph{Joint GRPO Optimization.}
Frame-selection and coordinate-conditioned grounding instructions are jointly sampled within each batch during reinforcement fine-tuning.
The reward for the $i$-th response is
\begin{equation}
R^{(i)}
=
R_{\mathrm{fmt}}^{(i)}
+
\begin{cases}
R_{\mathrm{grd}}^{(i)}, & \text{for grounding},\\
R_{\mathrm{sel}}^{(i)}, & \text{for frame selection}.
\end{cases}
\end{equation}
Through direct and indirect IoU rewards, both tasks are aligned with accurate 3D grounding.

\begin{table}[t]
\resizebox{\columnwidth}{!}{
\begin{tabular}{c|cc}
\hline
\rowcolor[HTML]{D9D9D9} 
\cellcolor[HTML]{D9D9D9}                         & \multicolumn{2}{c}{\cellcolor[HTML]{D9D9D9}ScanRefer} \\
\rowcolor[HTML]{D9D9D9} 
\multirow{-2}{*}{\cellcolor[HTML]{D9D9D9}Method} & $\mathrm{Acc}@0.25$                     & $\mathrm{Acc}@0.5$                      \\ \hline
ScanRefer                                        &                              &                              \\
\cite{chen2020scanrefer}        & \multirow{-2}{*}{37.3}       & \multirow{-2}{*}{24.3}       \\
\rowcolor[HTML]{F2F2F2} 
3D-LLM                                           & \cellcolor[HTML]{F2F2F2}     & \cellcolor[HTML]{F2F2F2}     \\
\rowcolor[HTML]{F2F2F2} 
\cite{hong20233d}   & \multirow{-2}{*}{\cellcolor[HTML]{F2F2F2}30.3} & \multirow{-2}{*}{\cellcolor[HTML]{F2F2F2}-}     \\
Grounded 3D-LLM                                  &                              &                              \\
\cite{chen2024grounded3dllm}    & \multirow{-2}{*}{47.9}       & \multirow{-2}{*}{44.1}       \\
\rowcolor[HTML]{F2F2F2} 
LLaVA-3D                                         & \cellcolor[HTML]{F2F2F2}     & \cellcolor[HTML]{F2F2F2}     \\
\rowcolor[HTML]{F2F2F2} 
\cite{zhu2024llava} & \multirow{-2}{*}{\cellcolor[HTML]{F2F2F2}54.1} & \multirow{-2}{*}{\cellcolor[HTML]{F2F2F2}42.4} \\
Video-3D LLM                                     &                              &                              \\
\cite{zheng2025video}           & \multirow{-2}{*}{58.1}       & \multirow{-2}{*}{51.7}       \\
\rowcolor[HTML]{F2F2F2} 
\textbf{Ours-2B$^{*}$}
& \textbf{60.0}
& \textbf{53.4} \\ \hline                        
\end{tabular}}
\vspace{-1.5mm}
\caption{{In-domain comparison with methods using explicit 3D input on ScanRefer.}
Our 3D-refined variant achieves the best performance among methods using explicit 3D information.
$^{*}$ denotes 3D refinement.}
\label{tab:my-table}
\vspace{-4mm}
\end{table}
\section{Experiment}
\subsection{Implementation details.}
We instantiate CoordRefer with Qwen3-VL-2B~\cite{bai2025qwen3} using a simple RGB-only architecture, without introducing additional geometry encoders such as VGGT~\cite{wang2025vggt}.
This minimal design allows us to isolate the effectiveness of the proposed coordinate-aware formulation from the benefits of explicit geometric representations or specialized architectural components.
During SFT, we train our model for one epoch on a mixed ScanNet-based dataset following VG-LLM~\cite{zheng2025learning}, covering detection, captioning, and 3D visual grounding, which consists of coordinate-frame selection and coordinate-conditioned grounding. RL is applied only to the two grounding subtasks.
At each training step, a mini-batch is randomly sampled from one of these tasks.
The model is optimized using Adam~\cite{kingma2014adam} with a learning rate of $1\times10^{-5}$.
During training, the visual encoder is frozen, while the MLLM backbone remains trainable.
For coordinate supervision, we use visible object annotations from EmbodiedScan~\cite{wang2024embodiedscan} together with ScanNet annotations. All SFT experiments are conducted on 16 NPUs, each with 64 GB of memory, using a total batch size of 16. RL experiments are performed on 64 such NPUs with the same batch size, generating 8 responses per input for single-object grounding and 4 responses per input for multi-object grounding.

\begin{figure*}[t]
  \centering
  \includegraphics[width=1\linewidth]
    {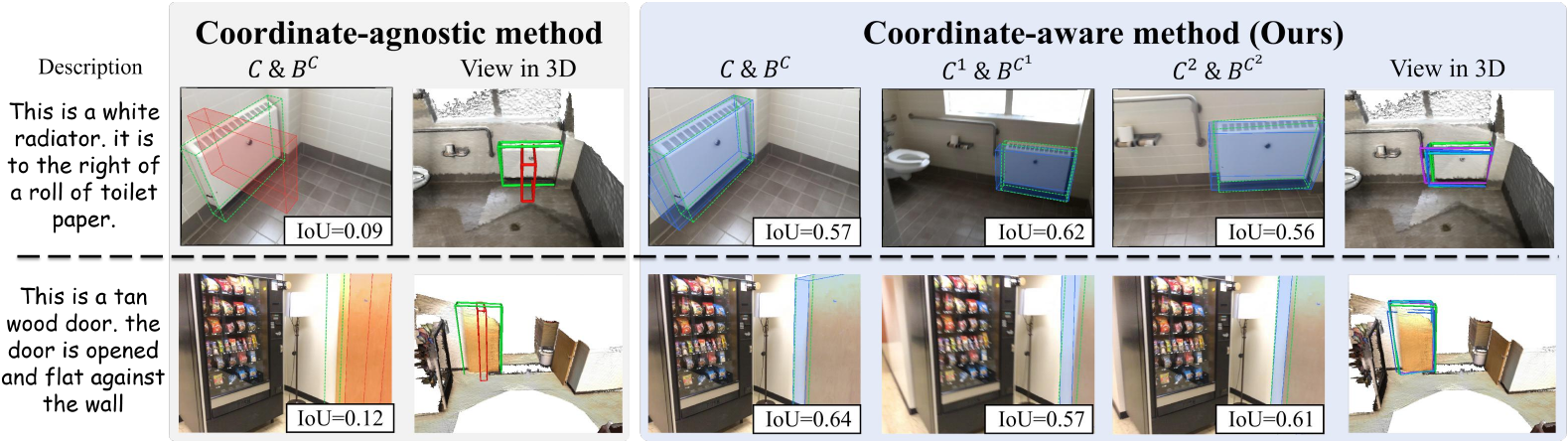}
    \vspace{-6mm}
   \caption{Qualitative evidence of resolving coordinate-relative box ambiguity.
The coordinate-agnostic baseline predicts an inconsistent coordinate--box pair, producing a box misaligned with the target in 3D space.
In contrast, our method predicts box representations under coordinates \(C\), \(C^{1}\), and \(C^{2}\), which align with the same object after transformation to a common 3D space.
Red, blue, and green denote coordinate-agnostic predictions, coordinate-aware predictions, and ground truth, respectively.
} 
\vspace{-4mm}
    \label{fig:ambiguity_vis}
\end{figure*}
\subsection{Single-object 3D Visual Grounding}
\label{sec:Single-Object-Grounding}

\paragraph{\textbf{Settings.}}

To evaluate single-object 3D visual grounding, we adopt ScanRefer~\cite{chen2020scanrefer}, NR3D and SR3D~\cite{achlioptas2020referit_3d}, and ARKitSceneRefer~\cite{kato2023arkitscenerefer} as benchmark datasets. We train and evaluate the model on ScanRefer, NR3D, and SR3D for in-domain evaluation. For zero-shot out-of-domain evaluation, we test the ScanRefer-trained model on the validation split of ARKitSceneRefer without additional fine-tuning.
ARKitSceneRefer is constructed from the ARKitScenes~\cite{dehghan2021arkitscenes} dataset. It therefore represents a more challenging cross-dataset setting, involving simultaneous shifts in scene geometry, image resolutions, camera intrinsics and extrinsics, object categories and scales. In particular, many targets occupy only small image regions, making both semantic identification and precise 3D box localization more difficult. 
We report $\mathrm{Acc}@0.25$ and $\mathrm{Acc}@0.5$, corresponding to localization accuracy at 3D IoU thresholds of 0.25 and 0.5. 
For each query, we select one reference frame, predict a 3D box in its camera coordinate system, transform it to the global coordinate system using the camera extrinsics, and compute its 3D IoU with the ground truth.
\paragraph{\textbf{Comparison with State-of-the-Art Methods.}}
As shown in Table~\ref{tab:RGB_method}, our RGB-only model achieves 51.1\% $\mathrm{Acc}@0.25$ and 23.6\% $\mathrm{Acc}@0.5$ on ScanRefer, substantially outperforming all prior methods that rely solely on multiview RGB images. Compared with the original Qwen3-VL-2B baseline trained under the conventional coordinate-agnostic formulation, our method yields absolute improvements of 11.4\% and 7.2\% at the two IoU thresholds, respectively. Since our model uses the same backbone without introducing additional geometric encoders, these gains cannot be attributed to model scaling or increased architectural capacity. Instead, they validate the effectiveness of our coordinate-aware method. 

\begin{figure*}[t]
  \centering
  \includegraphics[width=1\linewidth]
    {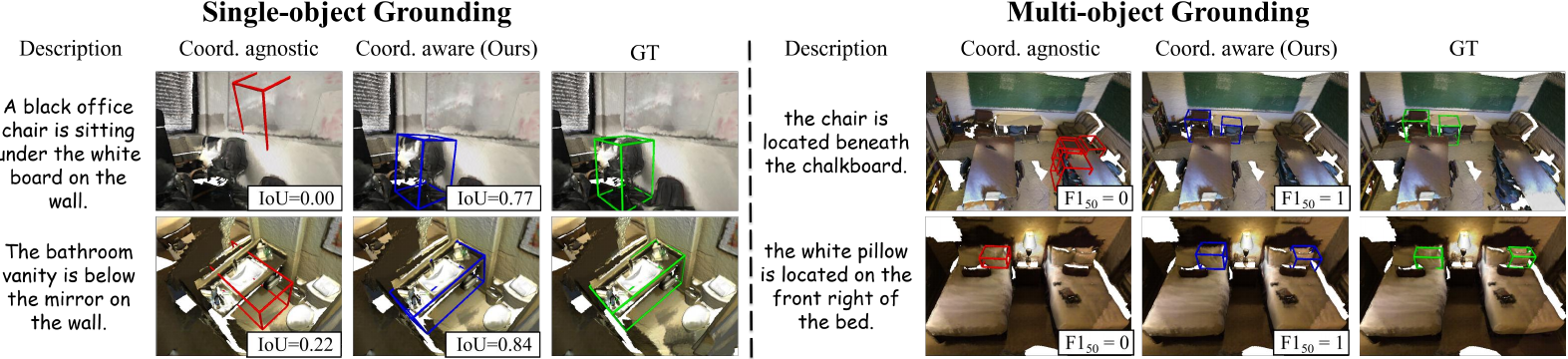}
    \vspace{-5mm}
    \caption{Qualitative results on single- and multi-object 3D visual grounding.
For single-object grounding, our method achieves more precise localization. For multi-object grounding, it accurately identifies both the number and locations of referred objects.
}
    \label{fig:single-multi}
    \vspace{-4mm}
\end{figure*}

Fig.~\ref{fig:ambiguity_vis} illustrates coordinate-relative box ambiguity and how our method resolves it.
Under the coordinate-agnostic formulation, the coordinate \(C\) and box \(B^{C}\) suffer from coordinate-relative box ambiguity. Consequently, the predicted box is misaligned with the ground-truth box in position and rotation, indicating that the predicted box is inconsistent with its associated coordinate frame.
In contrast, our method not only produces an accurate box under the selected coordinate \(C\), but also predicts different numerical representations \(B^{C^{1}}\) and \(B^{C^{2}}\) when conditioned on coordinates \(C^{1}\) and \(C^{2}\).
After transformation into the global 3D space, our coordinate-aware predictions consistently converge on the same physical object.
This behavior demonstrates that our framework establishes a stable correspondence between each coordinate condition and its box representation, thereby avoiding invalid compromise predictions caused by multiple equivalent optimization targets.
Fig.~\ref{fig:single-multi} provides further visual evidence of our model's ability to localize target objects.

For a fair comparison with methods that use explicit 3D information, we further apply the geometric refinement~\cite{zhang2025from} to the bounding boxes predicted by the VLM. Under this setting, our method achieves 60.0\% $\mathrm{Acc}@0.25$ and 53.4\% $\mathrm{Acc}@0.5$, outperforming Video-3D LLM~\cite{zheng2025video} by 1.9\% and 1.7\%, respectively. This result demonstrates that our framework remains highly competitive under both RGB-only and 3D-refined settings.

\paragraph{\textbf{Zero-Shot Out-of-Domain Generalization.}}

\begin{table}[t]
\centering
\setlength{\tabcolsep}{8pt}
\renewcommand{\arraystretch}{1.05}

\resizebox{\columnwidth}{!}{
\begin{tabular}{c|cc}
\hline
\rowcolor[HTML]{D9D9D9}
\cellcolor[HTML]{D9D9D9}
&
\multicolumn{2}{c}{\cellcolor[HTML]{D9D9D9}ARKitSceneRefer}
\\

\rowcolor[HTML]{D9D9D9}
\multirow{-2}{*}{\cellcolor[HTML]{D9D9D9}Method}
&
$\mathrm{Acc}@0.25$
&
$\mathrm{Acc}@0.5$
\\ \hline

VG-LLM-4B
&
&
\\
~\cite{zheng2025learning}
&
\multirow{-2}{*}{3.4}
&
\multirow{-2}{*}{0.2}
\\

\rowcolor[HTML]{F2F2F2}
~VG-LLM-8B
&
\cellcolor[HTML]{F2F2F2}
&
\cellcolor[HTML]{F2F2F2}
\\
\rowcolor[HTML]{F2F2F2}
~\cite{zheng2025learning}
&
\multirow{-2}{*}{\cellcolor[HTML]{F2F2F2}3.7}
&
\multirow{-2}{*}{\cellcolor[HTML]{F2F2F2}0.1}
\\

Qwen3-VL-2B$^\dagger$
&
&
\\
~\cite{bai2025qwen3}
&
\multirow{-2}{*}{6.2}
&
\multirow{-2}{*}{0.6}
\\

\rowcolor[HTML]{F2F2F2}
\textbf{Ours-2B}
&
\textbf{7.9}
&
\textbf{0.7}
\\ \hline

Video-3D LLM
&
&
\\
~\cite{zheng2025video}
&
\multirow{-2}{*}{0.2}
&
\multirow{-2}{*}{0.0}
\\ \hline

\end{tabular}
}
\vspace{-1.5mm}
\caption{Out-of-domain comparison on ARKitSceneRefer.
All models are trained on ScanRefer and evaluated without target-domain fine-tuning.
Our method achieves the best out-of-domain performance among all compared methods, demonstrating superior generalization over both RGB-based and explicit 3D-input approaches.$^\dagger$ denotes the coordinate-agnostic baseline instantiated with Qwen3-VL-2B and trained under the same experimental setting.}
\vspace{-4mm}
\label{tab:domain-out}
\end{table}
Table~\ref{tab:domain-out} evaluates the zero-shot generalization on ARKitSceneRefer of models trained only on ScanRefer. 
Under this particularly challenging cross-dataset setting, our RGB-only model achieves 7.9\% $\mathrm{Acc}@0.25$ and 0.7\% $\mathrm{Acc}@0.5$, outperforming both its Qwen3-VL-2B baseline and previous RGB-based methods. Although all approaches experience substantial performance degradation, our method retains the strongest grounding capability, demonstrating superior robustness to unseen scene distributions and acquisition conditions. 

In contrast, Video-3D LLM obtains only 0.2\% $\mathrm{Acc}@0.25$ and 0.0\% $\mathrm{Acc}@0.5$, exhibiting almost no cross-dataset generalization to ARKitSceneRefer. Since it relies on point-cloud-based proposal extraction, its performance is affected by the data gap between ScanNet and ARKitScenes, which may prevent the proposal module from generating reliable candidate boxes. These results highlight the cross-dataset robustness of our image-based, coordinate-aware grounding formulation.

\subsection{Multi-object 3D Visual Grounding}
\label{sec:Multiple-Object-Grounding}

\paragraph{\textbf{Settings.}}
To evaluate multi-object 3D visual grounding, we train and evaluate our method on Multi3DRefer~\cite{zhang2023multi3drefer}, which extends 3D visual grounding to a variable number of target objects.
Multi3DRefer is also built on ScanNet~\cite{dai2017scannet}.
Given a 3D scene and a referring query, the model is required to predict all 3D bounding boxes that satisfy the description.
We report F1 scores at IoU thresholds of 0.25 and 0.5 as the evaluation metrics.
For multi-object grounding, the model selects multiple reference frames and aggregates the resulting coordinate-conditioned predictions in the global coordinate system. During RL, F1 over the aggregated boxes rewards accurate localization while penalizing duplicate predictions. Formal definitions are provided in the supplementary material.
\paragraph{\textbf{Comparison on Multi-object Visual Grounding.}}
Table~\ref{tab:RGB_method} presents the results on Multi3DRefer. Our method achieves 45.4\% $\mathrm{F1}@0.25$ and 20.8\% $\mathrm{F1}@0.5$, outperforming all RGB-based methods. Compared with the Qwen3-VL-2B baseline, it yields absolute improvements of 11.0\% and 6.0\%, respectively, demonstrating the effectiveness of our framework for grounding a variable number of target objects.
Fig.~\ref{fig:single-multi} provides visual evidence that our model better captures the spatial distribution of objects in 3D space and more reliably localizes all referred instances in multi-object settings.

\begin{table}[t]
\centering

\setlength{\tabcolsep}{4.5pt}
\renewcommand{\arraystretch}{1.08}

\resizebox{\columnwidth}{!}{
\begin{tabular}{ccc|cc}
\hline

\rowcolor[HTML]{D9D9D9}
\multicolumn{3}{c|}{\cellcolor[HTML]{D9D9D9}SFT Components}
&
\multicolumn{2}{c}{\cellcolor[HTML]{D9D9D9}ScanRefer}
\\

\rowcolor[HTML]{D9D9D9}
Coord.
&
Suboptimal
&
Temp.
&
$\mathrm{Acc}@0.25$
&
$\mathrm{Acc}@0.5$
\\ \hline

-- & -- & -- 
& 39.7 & 16.4 \\

\ding{51} & -- & -- 
& 43.4 & 17.4 \\

\ding{51} & \ding{51} & -- 
& 44.4 & 18.7 \\

\ding{51} & \ding{51} & \ding{51}
& \textbf{47.1} & \textbf{20.3} \\ \hline

\end{tabular}
}
\vspace{-1.5mm}
\caption{{Ablation study on SFT components.}
Coord., Suboptimal, and Temp. denote the coordinate-aware formulation,
suboptimal-frame supervision, and temporal resampling.
Combining all components achieves the best performance.}
\vspace{-4.5mm}
\label{tab:sft_ablation}
\end{table}
\subsection{Ablation Study}
\label{sec:Ablation-Study}

\paragraph{\textbf{Ablation on SFT Components.}}
As shown in Table~\ref{tab:sft_ablation}, decoupling the two tasks by first predicting a reference-frame index and then grounding from only the selected image already clearly improves over the coordinate-agnostic baseline, validating our coordinate-aware formulation.
Suboptimal-Frame supervision further improves performance by enabling the grounding model to handle imperfect frame selections, while temporal resampling increases  the visual evidence in grounding. Combining these components achieves the best performance, demonstrating their complementary effects.

\paragraph{\textbf{Ablation on RL Components.}}
Table~\ref{tab:rl_ablation} studies the contributions of different RL components.
Simply continuing SFT beyond convergence does not improve performance, indicating that the gains are not due to additional training iterations. Applying an IoU-based reward to grounding produces clear improvements by directly aligning optimization with localization quality. For frame selection, the pseudo-label reward provides only limited benefits, whereas our indirect IoU reward achieves the best performance by evaluating each selected frame according to the grounding quality it enables. 

\begin{table}[t]
\centering

\setlength{\tabcolsep}{4pt}
\renewcommand{\arraystretch}{1.05}

\resizebox{\columnwidth}{!}{
\begin{tabular}{c|cc|cc}
\hline
\rowcolor[HTML]{D9D9D9}
&
\multicolumn{2}{c|}{\cellcolor[HTML]{D9D9D9}RL Reward}
&
\multicolumn{2}{c}{\cellcolor[HTML]{D9D9D9}ScanRefer}
\\

\rowcolor[HTML]{D9D9D9}
\multirow{-2}{*}{\cellcolor[HTML]{D9D9D9}Training}
&
Grounding
&
Coordinate
&
$\mathrm{Acc}@0.25$
&
$\mathrm{Acc}@0.5$
\\ \hline

SFT
& -- & --
& 47.1 & 20.3 \\

SFT + SFT
& -- & --
& 46.2 & 19.5 \\

SFT + RL
& IoU & --
& 50.7 & 22.6 \\

SFT + RL
& IoU & IoU
& \textbf{51.1}
& \textbf{23.6} \\ \hline

\textcolor{gray!60}{SFT + RL}
& \textcolor{gray!60}{IoU}
& \textcolor{gray!60}{PL}
& \textcolor{gray!60}{50.1}
& \textcolor{gray!60}{23.1} \\ \hline

\end{tabular}
}
\vspace{-1.5mm}
\caption{Ablation study on RL components.
Grounding and Coordinate denote the rewards for coordinate-conditioned
grounding and coordinate selection.
PL denotes the pseudo-label frame reward in 3D-RFT~\cite{linghu20263d}.
Using IoU rewards for both tasks achieves the best performance.}
\label{tab:rl_ablation}
\vspace{-4.5mm}
\end{table}
\section{Conclusion}
We revisit image-based 3D visual grounding and identify a structural limitation in existing frameworks, termed coordinate-relative box ambiguity. Jointly predicting the coordinate frame and 3D box introduces multiple equivalent optimization targets and may lead to invalid compromise predictions. To address this issue, we propose CoordRefer, a coordinate-aware framework that decouples coordinate-frame selection from 3D box regression and explicitly conditions grounding on the selected frame, yielding a unique and stable optimization target. We realize this formulation through a progressive training strategy, where coordinate-aware SFT first establishes explicit coordinate–box correspondence, followed by GRPO that aligns both grounding stages with localization quality through direct and indirect IoU rewards. Extensive experiments demonstrate consistent improvements across single- and multi-object grounding benchmarks, together with strong zero-shot cross-dataset generalization. Our RGB-only model substantially outperforms existing image-based methods, while its geometrically refined variant further surpasses approaches using explicit 3D inputs. We hope this study inspires future research on spatial reasoning and RGB-based 3D perception in MLLMs.
\clearpage
\appendix

\twocolumn[
\begin{center}
    {\LARGE\bfseries Supplementary Material}
\end{center}
\vspace{1em}
]

\section{Additional Experimental Details}
\label{sec:additional_details}

For \textbf{Coordinate-Aware Supervised Fine-Tuning}, we uniformly sample 32 images at equal temporal intervals from each scene to form $\mathcal{I}$ for coordinate frame selection, ensuring that the target object is visible in at least one frame. We construct 64K training samples for this stage. 
For coordinate-conditioned grounding, the selected reference frame is
placed first, followed by 18 temporally adjacent frames, forming
$\mathcal{I}_{\mathrm{grd}}$.
We construct 64K grounding samples using the optimal pseudo coordinate frames. To enhance robustness to imperfect coordinate selection, we further construct 64K grounding samples conditioned on valid suboptimal coordinate frames, resulting in 128K grounding samples in total.

For \textbf{Coordinate-Aware Reinforcement Learning}, all 64K coordinate frame selection samples and 128K coordinate-conditioned grounding samples are used for optimization.  We set the sampling temperature to 0.8, the $\mathrm{top}\_p$ to 0.95, and $\mathrm{top}\_k$ to 50. Following the DAPO~\cite{yu2026dapo}, we set $\epsilon_{\mathrm{high}}$ to 0.28, $\epsilon$ to 0.2, while keeping all other hyperparameters at the default settings of TRL's GRPOTrainer.

\paragraph{Software Environment and Reproducibility.}
All experiments are conducted with Python 3.10, PyTorch 2.6.0, and TRL 0.29.0. We use a random seed of 0 for all experiments.

\section{Definition of the Pseudo Coordinate-Frame Label}
\label{sec:pseudo_frame_selection}

A reliable pseudo coordinate-frame label should provide sufficient
visual and geometric evidence for downstream grounding.
We first use object visibility as a hard constraint to construct the
candidate-frame set. Among the visible candidate frames, projected
scale and geometric completeness are used to select the pseudo
coordinate-frame label.

\paragraph{Candidate-Frame Construction.}
Let $v_t \in \{0,1\}$ indicate whether the referred object is visible
in frame $t$. We define the candidate-frame set as
\begin{equation}
\mathcal{V}
=
\left\{
t \in \{1,\ldots,T\}
\mid v_t = 1
\right\},
\label{eq:visible_candidate_set}
\end{equation}
where $v_t$ is obtained from the visible-object annotations from EmbodiedScan~\cite{wang2024embodiedscan}.
Frames in which the target object is not visible are excluded from
subsequent scoring and cannot be selected as coordinate frames.

\begin{figure*}[t]
  \centering
  \includegraphics[width=1\linewidth]
    {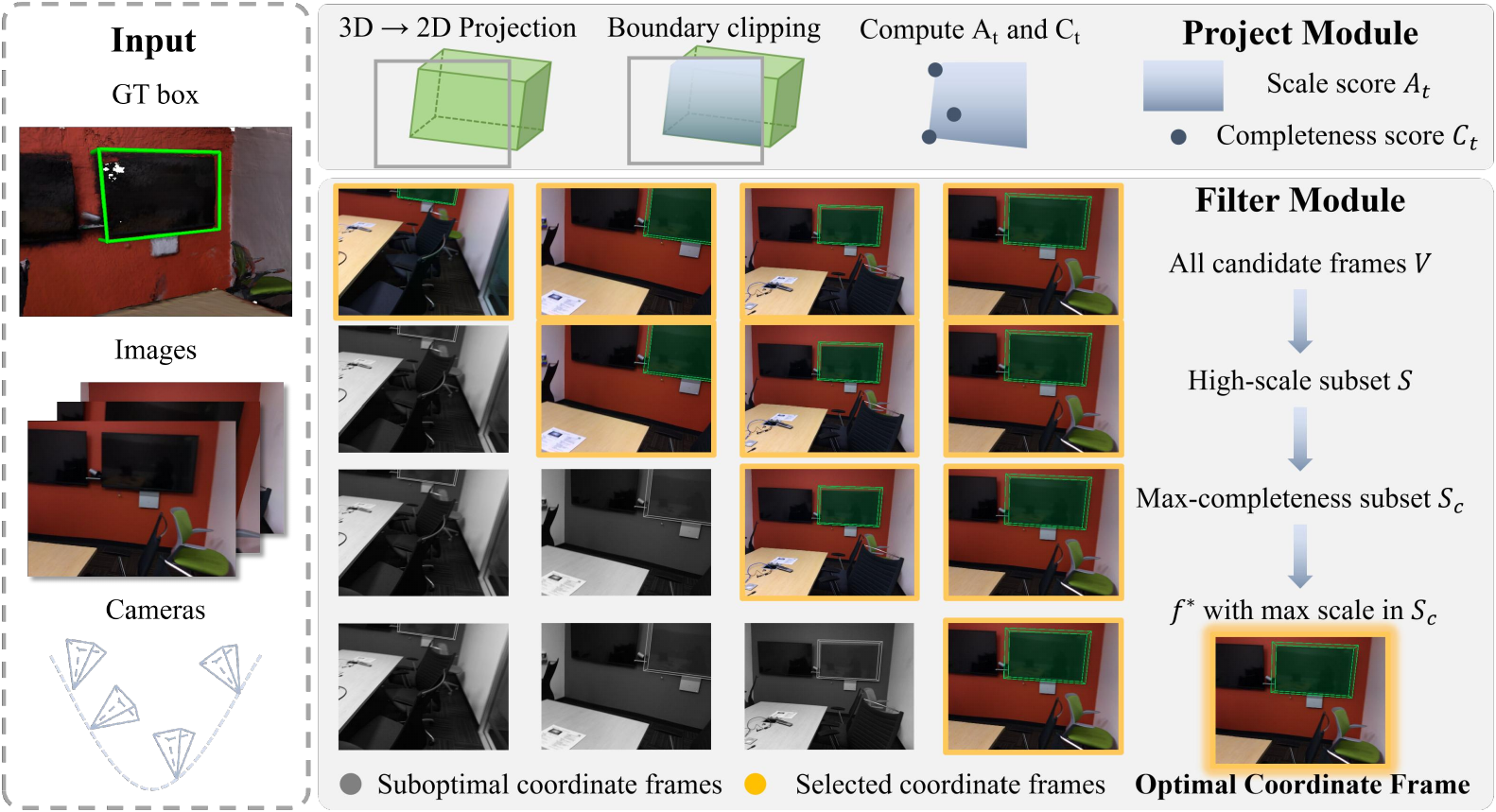}
  \caption{
    {Pseudo coordinate-frame label.}
    We first exclude frames in which the target object is not visible.
    For each remaining candidate frame, we project the 3D bounding box
    into the image plane and compute its projected area $A_t$ and
    geometric completeness $C_t$ after boundary clipping.
    The filter module then selects the pseudo coordinate frame by
    retaining sufficiently large projections and prioritizing geometric
    completeness.
  }
  \label{fig:frameselection}
\end{figure*}

\paragraph{Scale--Completeness Balanced Selection.}
As shown in Fig.~\ref{fig:frameselection}, for each candidate frame
$t \in \mathcal{V}$, we project the eight corners of the 3D bounding
box from world coordinates onto the image plane using the camera
extrinsic and intrinsic matrices.
Only corners with positive depth in the camera coordinate system are
retained, and the projected coordinates are clipped to the image
boundaries.
We compute two scores for each candidate frame:
\begin{itemize}
    \item \textbf{Scale score $A_t$.}
The scale score $A_t$ is defined as the normalized projected 2D area after clipping. It measures the amount of geometric evidence contained in the image.
    \item \textbf{Completeness score $C_t$.}
The completeness score $C_t$ is defined using the corners that are both in front of the camera and within the image boundaries. It measures how fully the object is visible in the image. 
\end{itemize}
We then construct a high-scale subset:
\begin{equation}
\mathcal{S} = \left\{ t \in \mathcal{V} \ \big| \ A_t \ge \tau \cdot \max_{j \in \mathcal{V}} A_j \right\},
\end{equation}
where $\tau \in [0,1]$ is a fixed ratio threshold used to exclude frames with insufficient projection scale. 
Since the frame achieving $\max A_j$ always satisfies the condition, $\mathcal{S}$ is guaranteed to be non-empty.

The final coordinate frame is selected from $\mathcal{S}$ with completeness given higher priority:
\begin{equation}
\mathcal{S}_C = 
\left\{ 
t \in \mathcal{S} \mid 
C_t = \max_{k \in \mathcal{S}} C_k 
\right\}
\end{equation}

\begin{equation}
f^* = \arg\max_{t \in \mathcal{S}_C} A_t ,
\label{equ:frame_choose}
\end{equation}

By balancing scale and completeness, we avoid selecting coordinate frames that are either truncated or too small, providing a more stable geometric reference for grounding.
The balancing threshold $\tau$ is empirically set to 0.6. The ablation study of $\tau$ is reported in Table~\ref{tab:tau_ablation}. We observe that $\tau = 0.6$ provides the best overall trade-off across the two IoU thresholds. Note that when $\tau = 0$, the definition prioritizes completeness over all candidate frames, while when $\tau = 1$, it is dominated by projected scale. This ablation is conducted at the SFT stage, before reinforcement learning is introduced. 

\begin{table}[t]
\centering
\small
\setlength{\tabcolsep}{6pt}
\renewcommand{\arraystretch}{1.15}

\begin{tabular}{c|cc}
\hline
\rowcolor[HTML]{D9D9D9}
&
\multicolumn{2}{c}{\cellcolor[HTML]{D9D9D9}\textbf{ScanRefer}}
\\

\rowcolor[HTML]{D9D9D9}
\multirow{-2}{*}{\cellcolor[HTML]{D9D9D9}\textbf{$\tau$}}
&
\textbf{$\mathrm{Acc}@0.25$}
&
\textbf{$\mathrm{Acc}@0.5$}
\\
\hline

0   & 44.9 & 19.1 \\
0.2 & 45.3 & 20.0 \\
0.4 & 46.0 & 19.8 \\
0.6 & \textbf{47.1} & \underline{20.3} \\
0.8 & 45.8 & \textbf{20.5} \\
1   & \underline{46.5} & 19.7 \\

\hline
\end{tabular}

\caption{{Ablation on threshold $\tau$ for coordinate selection.}
We vary the threshold $\tau$ and report grounding performance on ScanRefer.
The best results are highlighted in bold, and the second-best results are underlined.}
\label{tab:tau_ablation}
\end{table}

\section{Training Data Examples and Details for the Coordinate-Aware Framework}
\label{sec:training_data}
We provide the prompt templates for single- and multi-object 3D visual grounding in Tables~\ref{tab:prompt_single} and~\ref{tab:prompt_multiple}, respectively. In both settings, each selected reference frame is placed at the beginning of its grounding image sequence, followed by the images obtained through temporal resampling. The prompt instructs the model to express its 3D predictions in the camera coordinate system of the first frame, which serves as the reference frame.

\begin{table*}[!t]
\centering

\begin{tcolorbox}[
  enhanced,
  width=\linewidth,
  colback=gray!10,
  colframe=black!70,
  boxrule=0.8pt,
  arc=3.5mm,
  left=2.5mm,right=2.5mm,top=2mm,bottom=2mm,
  boxsep=1.2mm,
  before upper=\raggedright,
]

\tasktitle{Single-Object 3D Visual Grounding -- Coordinate Frame Selection}

\Xuser\par
\frameitem{0}\;\frameitem{1}\;\frameitem{2}\;\frameitem{3}\;$\dots$\par
\smallskip
\textbf{Text: There is a beige wooden bookshelf placed next to another bookshelf.}\par
Localize the clearest frame in the video showing the object described in the text.
Output a JSON dictionary with a frame index in \texttt{"frame"}.\par
\smallskip
\Xagent\par
\begin{lstlisting}[style=jsonstyle]
(*@\fencejson@*)
{"frame": 1}
(*@\fence@*)
\end{lstlisting}

\tcblower

\tasktitle{Single-Object 3D Visual Grounding -- Coordinate-Conditioned Grounding}

\Xuser\par
\frameitem{0}\;\frameitem{1}\;\frameitem{2}\; \frameitem{3}\;$\dots$\par
\smallskip

\textbf{Text: There is a beige wooden bookshelf placed next to another bookshelf.}\par
Output a JSON dictionary that matches the description and includes the 3D bounding box of the object,
with the bounding box data stored in \texttt{"bbox\_3d"}, where the coordinates are based on the coordinate
system of the first frame. The 3D bounding box format should be
\texttt{[x\_center, y\_center, z\_center, x\_size, y\_size, z\_size, yaw, pitch, roll]}.\par

\smallskip
\Xagent\par
\begin{lstlisting}[style=jsonstyle]
(*@\fencejson@*)
{"bbox_3d": [-0.63, -0.83, 2.43, 3.0, 0.59, 2.35, -2.32, 1.18, 3.05]}
(*@\fence@*)
\end{lstlisting}

\end{tcolorbox}
\caption{{The prompts for Single-Object 3D visual grounding tasks.} The output should be in JSON format.}
\label{tab:prompt_single}
\end{table*}

\begin{table*}[!t]
\centering
\begin{tcolorbox}[
  enhanced,
  width=\linewidth,
  colback=gray!10,
  colframe=black!70,
  boxrule=0.8pt,
  arc=3.5mm,
  left=2.5mm,right=2.5mm,top=2mm,bottom=2mm,
  boxsep=1.2mm,
  before upper=\raggedright,
]

\tasktitle{Multi-Object 3D Visual Grounding -- Coordinate Frame Selection}

\Xuser\par
\frameitem{0}\;\frameitem{1}\;\frameitem{2}\;
\frameitem{3}\;$\dots$\par
\smallskip

\textbf{Text: the small white nightstand is situated against the wall.}  \par
Localize the clearest frames in the video showing the objects described in the text. There may be zero, one, or multiple objects matching the text. Output a JSON dictionary whose "frame" field contains a duplicate-free list of selected frame indices\par

\smallskip
\Xagent\par
\begin{lstlisting}[style=jsonstyle]
(*@\fencejson@*)
{"frame": [28, 30]}
(*@\fence@*)
\end{lstlisting}

\tcblower

\tasktitle{Multi-Object 3D Visual Grounding -- Coordinate-Conditioned Grounding}

\Xuser\par
\frameitem{0}\;\frameitem{1}\;\frameitem{2}\;
\frameitem{3}\;$\dots$\par
\smallskip

\textbf{Text: the small white nightstand is situated against the wall.}\par
Detect all objects in the video that match the textual description.
For each detected object, output its 3D bounding box in the camera coordinate system of the first frame.
There may be zero, one, or multiple objects matching the text.
Output a JSON list where each entry contains its 3D bounding box in "bbox\_3d" in the first frame's coordinate.
The 3D bounding box format should be \texttt{[x\_center, y\_center, z\_center, x\_size, y\_size, z\_size, yaw, pitch, roll]}\par

\smallskip
\Xagent\par
\begin{lstlisting}[style=jsonstyle]
(*@\fencejson@*)
[
{"bbox_3d": [0.63, -0.83, 2.43, 3.0, 0.59, 2.35, 2.32, 1.18, 3.05]},
{"bbox_3d": [-0.87, 0.44, 3.53, 0.41, 0.48, 0.49, 0.73, 1.38, 2.92]}
]
(*@\fence@*)
\end{lstlisting}
\end{tcolorbox}
\caption{{The prompts for Multi-Object 3D visual grounding tasks.} The output should be in JSON format.}
\label{tab:prompt_multiple}
\end{table*}

\section{Extension to Multi-Object 3D Visual Grounding}
\label{sec:multi_object_extension}

\paragraph{Multi-Frame Coordinate-Aware Grounding.}
We extend the coordinate-aware formulation to multi-object 3D visual
grounding on Multi3DRefer~\cite{zhang2023multi3drefer}, where a textual query may refer to zero, one, or multiple
objects. Since the referred instances may not be sufficiently visible
in a single view, the coordinate-frame selection task predicts a
duplicate-free set of reference frames:
\begin{equation}
\hat{\mathcal{S}}_{f}
=
\mathcal{F}_{\mathrm{sel}}^{\mathrm{multi}}
(\mathcal{I},q)
=
\{\hat{f}_{k}\}_{k=1}^{K},
\label{eq:multi_frame_selection}
\end{equation}
where $K$ is determined by the model and may vary across queries.
For each selected frame $\hat{f}_{k}$, we construct a grounding input
$\mathcal{I}_{\mathrm{grd}}^{\hat{f}_{k}}$ using the same temporal
resampling strategy as in the single-object setting. The
coordinate-conditioned grounding model then predicts a set of 3D
bounding boxes in the coordinate system of $\hat{f}_{k}$:
\begin{equation}
\hat{\mathcal{B}}^{\hat{f}_{k}}
=
\mathcal{F}_{\mathrm{grd}}^{\mathrm{multi}}
\left(
\mathcal{I}_{\mathrm{grd}}^{\hat{f}_{k}},
q
\mid
\hat{f}_{k}
\right)
=
\left\{
\hat{B}_{k,j}^{\hat{f}_{k}}
\right\}_{j=1}^{N_k},
\label{eq:multi_conditioned_grounding}
\end{equation}
where $N_k$ is the number of boxes predicted under the coordinate
system of frame $\hat{f}_{k}$.
Predictions from all selected frames are transformed into a common
global coordinate system using the corresponding camera extrinsics and
aggregated into a single prediction set
$\hat{\mathcal{B}}^{g}$. We then match the aggregated predictions with
the ground-truth box set $\mathcal{B}^{*}$ following the
Multi3DRefer evaluation protocol.

\paragraph{Multi-Object SFT Target Construction.}
During supervised fine-tuning, we independently determine an optimal
pseudo coordinate frame for each referred object using
the same scale--completeness criterion as in the single-object setting.
The resulting frame indices are deduplicated to construct the
multi-object coordinate-frame selection target. For each retained
frame, the grounding target contains all referred instances visible in
that frame, with their ground-truth boxes transformed into its camera
coordinate system.
This construction improves the coverage of referred instances across
different viewpoints. However, the same physical instance may be
supervised under multiple coordinate frames and consequently predicted
multiple times after global aggregation.

\paragraph{F1-Based Reinforcement Fine-Tuning.}
To jointly optimize instance coverage and duplicate suppression, the
reward for the $i$-th multi-object response is defined as
\begin{equation}
R_{\mathrm{multi}}^{(i)}
=
R_{\mathrm{fmt}}^{(i)}
+
\operatorname{F1}_{0.25}
\left(
\hat{\mathcal{B}}^{g,(i)},
\mathcal{B}^{*}
\right)
+
\operatorname{F1}_{0.5}
\left(
\hat{\mathcal{B}}^{g,(i)},
\mathcal{B}^{*}
\right),
\label{eq:multi_object_reward}
\end{equation}
where $\operatorname{F1}_{0.25}$ and
$\operatorname{F1}_{0.5}$ denote the F1 scores computed at 3D IoU
thresholds of $0.25$ and $0.5$, respectively. Under the official
one-to-one matching procedure, missed instances decrease recall,
whereas duplicate predictions of the same physical instance are
treated as false positives and decrease precision. The F1-based reward
therefore encourages the model to cover all referred instances while
suppressing duplicate predictions across different grounding passes.

\paragraph{Reward Computation for the Two Tasks.}
For coordinate-conditioned grounding instructions,
$\hat{\mathcal{B}}^{g,(i)}$ is obtained by transforming and aggregating
the policy predictions. For coordinate-frame selection instructions,
the selected frame set is evaluated using the periodically refreshed
grounding reward model introduced in the single-object setting. The
reward model performs grounding under each selected frame, and its
predictions are transformed and aggregated to obtain
$\hat{\mathcal{B}}^{g,(i)}$.
The format reward requires a valid duplicate-free frame-index list for
coordinate-frame selection and a parseable list of complete 9-DoF
bounding boxes for coordinate-conditioned grounding. 

\section{Efficiency Comparison with the Coordinate-agnostic Baseline}
\label{sec:inference_time}
All experiments are conducted on ScanRefer with Qwen3-VL-2B~\cite{bai2025qwen3}. 
The training time and computational resources for the two optimization stages are shown in Table~\ref{tab:trainingtime}. 
\begin{table}[t]
\centering
\small
\setlength{\tabcolsep}{7pt}
\renewcommand{\arraystretch}{1.2}
\begin{tabular}{c c c}
\hline
\rowcolor[HTML]{D9D9D9}
\textbf{Training Stage}
& \textbf{NPU Configuration}
& \textbf{Time (h)} \\
\hline
SFT & $16 \times 64$GB NPUs & 20  \\
RL  & $64 \times 64$GB NPUs & 111 \\
\hline
\end{tabular}
\caption{{Training time and computational resources for SFT and RL.}
We report the total training time and the corresponding NPU configuration for each training stage.}
\label{tab:trainingtime}
\end{table}
The inference efficiency results are reported in Table~\ref{tab:efficiency_comparison}. 
Our framework increases the per-sample inference time from 8.2\,s to 10.4\,s, mainly due to the additional coordinate-frame selection stage. 
Nevertheless, the overall overhead is limited. 
Given the clear performance gains brought by disentangling coordinate frame selection and coordinate-conditioned grounding, we believe this additional cost is justified.

\begin{table}[t]
\centering

\small
\setlength{\tabcolsep}{8pt}
\renewcommand{\arraystretch}{1.2}
\begin{tabular}{c c c c}
\hline
\cellcolor[HTML]{D9D9D9} & \multicolumn{3}{c}{\cellcolor[HTML]{D9D9D9}\textbf{Inference per sample (s)}} \\
\rowcolor[HTML]{D9D9D9}
\multirow{-2}{*}{\textbf{Method}} & Frame & Ground & Total \\
\hline
Coord. agnostic & -- & -- & 8.2 \\
Coord. aware (Ours) & 4.1 & 6.3 & 10.4 \\
\hline
\end{tabular}
\caption{{Inference Efficiency.} Per-sample inference latency (s). Despite the additional stage, our coordinate-aware framework introduces only modest overhead.
}
\vspace{-4mm}
\label{tab:efficiency_comparison}
\end{table}

\section{Shared versus Task-Specific RL Policies}

Our default implementation uses a single 2B policy for both
coordinate-frame selection and coordinate-conditioned grounding. To
examine whether sharing the policy causes interference between the two
tasks during reinforcement learning, we additionally evaluate a
task-specific setting with two separate 2B policies.
In the task-specific setting, we first optimize the grounding policy using the direct IoU reward. After grounding optimization is completed, the resulting policy is frozen and used as a fixed grounding reward model for training the coordinate-frame selection policy. Unlike the default shared-policy setting, the reward model is not periodically refreshed during frame-selection training. This sequential design provides a stable reward signal, allowing the frame-selection policy to reliably learn the grounding model's preference over coordinate frames. Both policies are initialized from the same coordinate-aware SFT checkpoint and use the same task-specific training data and RL hyperparameters as the corresponding stages of the shared-policy setting.

As shown in Table~\ref{tab:shared_separate_policy}, separate training slightly improves $\mathrm{Acc}@0.25$ from 51.1\% to 51.9\%, while $\mathrm{Acc}@0.5$ marginally decreases from 23.6\% to 23.5\%. The improvements are therefore small and inconsistent across metrics, indicating that jointly optimizing the two tasks does not introduce substantial negative interference. More importantly, the shared-policy setting achieves comparable grounding performance using only a single 2B model, whereas separate training requires two task-specific 2B models. We therefore adopt the shared policy as the default setting due to its substantially lower model-storage and deployment overhead.
\begin{table}[t]
\centering
\setlength{\tabcolsep}{6pt}
\renewcommand{\arraystretch}{1.15}

\resizebox{\columnwidth}{!}{
\begin{tabular}{c|c|cc}
\hline

\rowcolor[HTML]{D9D9D9}

&

&
\multicolumn{2}{c}{\textbf{ScanRefer}}
\\

\rowcolor[HTML]{D9D9D9}\multirow{-2}{*}{\textbf{Policy Setting}}
&\multirow{-2}{*}{\textbf{Model Parm.}}
&
\textbf{$\mathrm{Acc}@0.25$}
&
\textbf{$\mathrm{Acc}@0.5$}
\\ \hline

Shared Policy
&
2B
&
51.1
&
\textbf{23.6}
\\

Task-Specific Policies
&
2B + 2B
&
\textbf{51.9}
&
23.5
\\ \hline

\end{tabular}
}

\caption{{Comparison of shared and task-specific RL policies.}
The shared-policy setting uses a single 2B model for both coordinate-frame
selection and coordinate-conditioned grounding, whereas the task-specific
setting uses two separate 2B models.
Task-specific policies provide only marginal and inconsistent performance
changes while doubling the model-storage and deployment requirements.}
\label{tab:shared_separate_policy}
\end{table}

\section{Reinforcement Learning Dynamics}
\label{sec:rl_dynamics}

\begin{figure}[t]
    \centering
    \includegraphics[width=\linewidth]
    {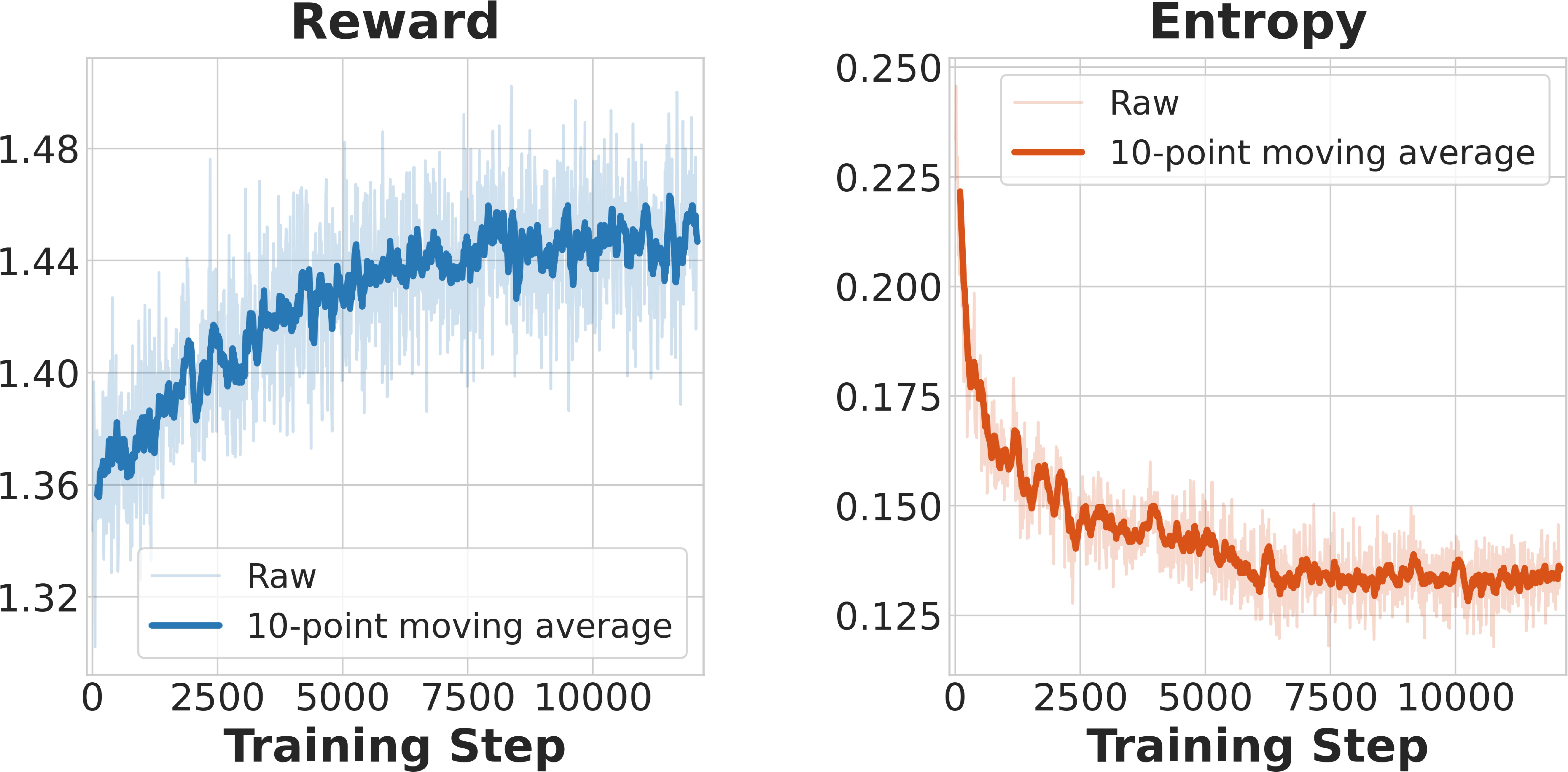}
    \caption{{Training dynamics of coordinate-aware reinforcement
    learning.}
    The left panel shows the training reward, while the right panel shows
    the policy entropy.}
    \label{fig:reward_entropy}
\end{figure}

To further analyze the optimization behavior of our coordinate-aware
reinforcement learning, we visualize the training reward and policy entropy
throughout the RL stage. As shown in Fig.~\ref{fig:reward_entropy},
the raw measurements exhibit moderate step-wise fluctuations due to stochastic
group sampling, while their 10-point moving averages reveal clear and stable
optimization trends. The average reward increases steadily from approximately
1.36 at the beginning of training to around 1.45, with the improvement gradually
saturating after about 8K steps. This trend indicates that the policy
progressively learns to generate coordinate-frame selections and grounding
predictions that are better aligned with the proposed IoU-based reward
objectives. Meanwhile, the policy entropy decreases rapidly during the early
training stage and subsequently stabilizes at approximately 0.13. The decreasing
entropy suggests that the policy becomes increasingly confident as optimization
proceeds, whereas its stable non-zero value indicates that the model retains
sufficient output diversity without evident policy collapse. Overall, the
steadily increasing reward and smoothly converging entropy demonstrate that the
proposed RL optimization is effective and remains stable throughout training.

\section{Upper-Bound Analysis of Coordinate Frame Modeling}
\label{sec:Upper_bound}

\begin{figure*}
  \centering
  \includegraphics[width=1\linewidth]
    {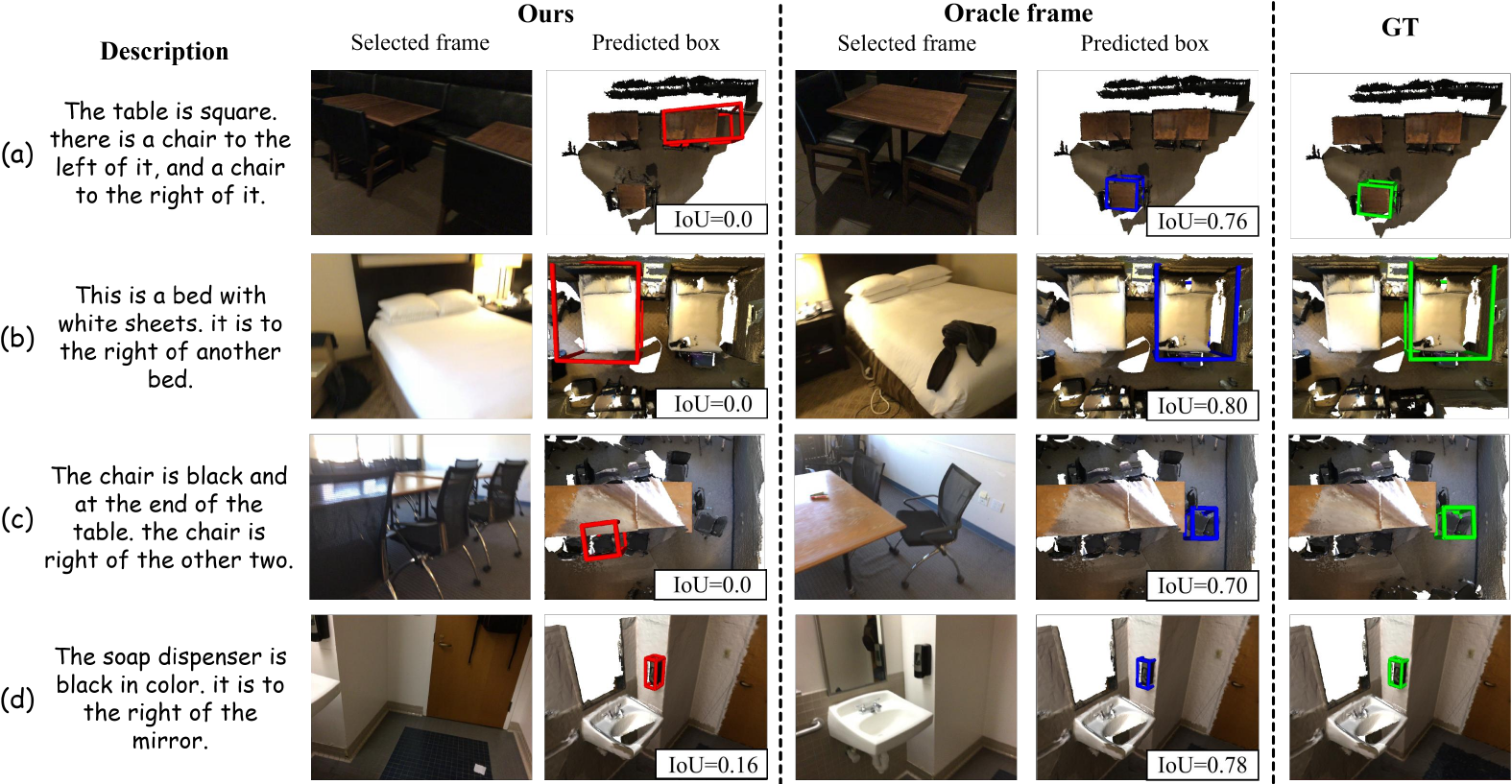}
    \caption{{Qualitative comparison between our two-stage framework and oracle coordinate frame selection.}
Compared with CoordRefer, the oracle frame often provides more informative viewpoints and leads to more accurate localization, revealing significant room for improvement in coordinate frame selection.}
    \label{fig:supply}
\end{figure*}

\begin{table}[t]
\centering

\setlength{\tabcolsep}{5pt}
\renewcommand{\arraystretch}{1.05}

\resizebox{\columnwidth}{!}{
\begin{tabular}{cc|cc}
\hline

\rowcolor[HTML]{D9D9D9}
\multicolumn{2}{c|}{\cellcolor[HTML]{D9D9D9}}
&
\multicolumn{2}{c}{\cellcolor[HTML]{D9D9D9}\textbf{ScanRefer}}
\\

\rowcolor[HTML]{D9D9D9}
\multirow{-2}{*}{\cellcolor[HTML]{D9D9D9}\textbf{Oracle Frame}}
&
\multirow{-2}{*}{\cellcolor[HTML]{D9D9D9}\textbf{Box Refinement}}
&
\textbf{$\mathrm{Acc}@0.25$}
&
\textbf{$\mathrm{Acc}@0.5$}
\\ \hline

--        & --        & 51.1 & 23.6 \\
--        & \ding{51} & 60.0 & 53.4 \\
\ding{51} & --        & 83.2 & 51.3 \\
\ding{51} & \ding{51} & \textbf{85.1} & \textbf{76.5} \\

\hline
\end{tabular}
}
\caption{{Upper-bound analysis on ScanRefer.}
Oracle Frame selects the coordinate frame yielding the highest grounding IoU, while Box Refinement denotes the geometric refinement used in our 3D-refined variant.}
\label{tab:upper_bound}
\end{table}

We conduct an upper-bound analysis on ScanRefer~\cite{chen2020scanrefer} using Qwen3-VL-2B~\cite{bai2025qwen3} to investigate the limitations of coordinate frame selection and 3D box regression.
For each sample, all input frames are treated as candidate coordinate frames, and the frame whose predicted box achieves the highest IoU with the ground truth is selected as the \emph{oracle frame}.
We further apply box refinement to improve the geometric accuracy of the predicted box.
As shown in Table~\ref{tab:upper_bound}, using the oracle frame improves $\mathrm{Acc}@0.25$/0.5 from 51.1\%/23.6\% to 83.2\%/51.3\%, demonstrating that coordinate frame selection remains a major performance bottleneck.
Box refinement~\cite{zhang2025from} improves the CoordRefer to 60.0\%/53.4\%, indicating that many predictions roughly localize the target but remain geometrically imprecise.
Combining the oracle frame and box refinement achieves the best performance of 85.1\%/76.5\%.
The results suggest that coordinate frame selection mainly determines whether the target can be correctly localized, while box refinement primarily improves fine-grained geometric accuracy.
Therefore, further improvements require both more reliable coordinate frame modeling and more precise coordinate-conditioned 3D box regression.

\begin{figure*}
\centering
\includegraphics[width=\linewidth]{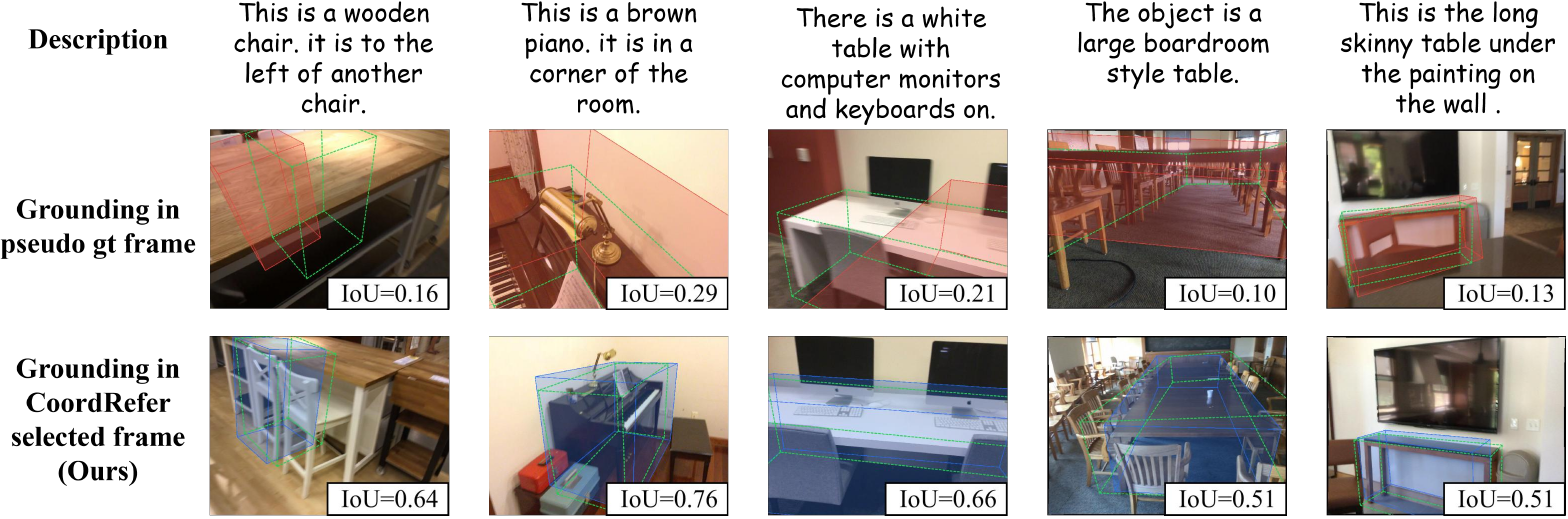}
\caption{{Qualitative comparison of heuristic and IoU-optimized coordinate-frame selection.}
The top and bottom rows show the coordinate frames selected by the handcrafted scale--completeness criterion and our indirect IoU reward, respectively.
IoU-based optimization selects frames that better support downstream grounding and consequently produces more accurate 3D bounding boxes.
Green boxes denote the ground truth, while red and blue boxes denote predictions based on heuristic and IoU-optimized coordinate frames, respectively.}
\label{fig:coordinate_selection_vis}
\end{figure*}

\section{The source of Camera Pose}
\label{sec:pose_estimation}

Our main experiments use ground-truth camera poses to transform predictions from different coordinate frames into a common 3D space, following the standard setting adopted by existing RGB-based methods.
To evaluate the dependence of our method on camera pose accuracy, we replace the ground-truth poses with poses estimated by $\pi^3$~\cite{wang2026pi} during inference.
As shown in Table~\ref{tab:pose_estimation}, using estimated poses decreases $\mathrm{Acc}@0.25$ from 51.1\% to 49.3\% and $\mathrm{Acc}@0.5$ from 23.6\% to 20.7\%.
Although estimated poses introduce a moderate performance drop, the model retains most of its grounding capability. These results indicate that CoordRefer remains effective when ground-truth poses are replaced with estimated poses.

\begin{table}[t]
\centering
\small
\setlength{\tabcolsep}{6pt}
\renewcommand{\arraystretch}{1.12}

\resizebox{\columnwidth}{!}{
\begin{tabular}{c|cc}
\hline
\rowcolor[HTML]{D9D9D9}
&
\multicolumn{2}{c}{\cellcolor[HTML]{D9D9D9}\textbf{ScanRefer}}
\\

\rowcolor[HTML]{D9D9D9}
\multirow{-2}{*}{\cellcolor[HTML]{D9D9D9}\textbf{Pose Source}}
&
\textbf{$\mathrm{Acc}@0.25$}
&
\textbf{$\mathrm{Acc}@0.5$}
\\
\hline

GT
& \textbf{51.1}
& \textbf{23.6}
\\

$\pi^3$~\cite{wang2026pi}
& 49.3
& 20.7
\\

\bottomrule
\end{tabular}
}

\caption{{Influence of camera pose estimation.}
Ground-truth poses correspond to the standard evaluation setting used in our main experiments and other RGB-based methods.}
\label{tab:pose_estimation}
\end{table}

\begin{table}[t]
\centering
\setlength{\tabcolsep}{4pt}
\renewcommand{\arraystretch}{1.15}

\resizebox{\columnwidth}{!}{
\begin{tabular}{c| c| c c c}
\hline
\rowcolor[HTML]{D9D9D9}
\textbf{Method}
& \textbf{Category}
& \makecell{\textbf{Precision}\\\textbf{@0.25}}
& \makecell{\textbf{Recall}\\\textbf{@0.25}}
& \makecell{\textbf{F1}\\\textbf{@0.25}} \\
\hline

\multirow{3}{*}{Coord. agnostic}
& cate8  & 45.2 & 41.2 & 43.1 \\
& cate20 & 38.6 & 32.7 & 35.2 \\
& cate31 & 30.5 & 26.0 & 27.9 \\

\hline

\multirow{3}{*}{Coord. aware (Ours)}
& cate8  & \textbf{49.0} & \textbf{43.6} & \textbf{46.1} \\
& cate20 & \textbf{41.0} & \textbf{34.7} & \textbf{37.3} \\
& cate31 & \textbf{32.2} & \textbf{27.6} & \textbf{29.4} \\

\hline
\end{tabular}
}

\caption{{Detection performance comparison at an IoU threshold of 0.25.}
Our method consistently outperforms the coordinate-agnostic framework across different category settings, demonstrating that the proposed two-stage design does not compromise detection capability.}

\label{tab:detection}
\end{table}
\begin{table}[t]
\centering
\setlength{\tabcolsep}{12pt}
\renewcommand{\arraystretch}{1.18}

\resizebox{\columnwidth}{!}{
\begin{tabular}{c|ccc}
\hline
\rowcolor[HTML]{D9D9D9}
&
\multicolumn{3}{c}{\cellcolor[HTML]{D9D9D9}\textbf{ScanCap}} \\
\rowcolor[HTML]{D9D9D9}
\multirow{-2}{*}{\cellcolor[HTML]{D9D9D9}\textbf{Method}}
& \textbf{C@0.5}
& \textbf{B-4@0.5}
& \textbf{R@0.5} \\
\hline

Coord. agnostic & \textbf{61.1} & 36.6 & 60.5 \\
Coord. aware (Ours)  & 60.7 & \textbf{36.8} & \textbf{60.8} \\

\bottomrule
\end{tabular}}

\caption{{Captioning performance comparison at an IoU threshold of 0.5.}
Our method achieves captioning performance comparable to the coordinate-agnostic framework,
demonstrating that the proposed design does not compromise captioning capability.}
\label{tab:caption}
\end{table}

\section{Impact on Detection and Captioning}
\label{sec:other_tasks}
We further evaluate whether the proposed framework affects other capabilities of the model following VG-LLM~\cite{zheng2025learning}, including 3D object detection and 3D captioning.
As shown in Table~\ref{tab:detection}, our method improves detection performance over the coordinate-agnostic baseline across all category settings. Precision, recall, and F1 scores improve consistently across the reported settings.

Similarly, Table~\ref{tab:caption} reports the captioning results. Our method achieves comparable CIDEr, BLEU-4, and ROUGE scores to the coordinate-agnostic baseline, indicating that the proposed design does not compromise captioning ability.
These results demonstrate that the performance gains in 3D visual grounding are achieved without sacrificing detection or captioning performance.

\section{Qualitative Analysis of Coordinate-Frame Selection}

Fig.~\ref{fig:coordinate_selection_vis} further illustrates the limitation of manually designed selection criteria.
The handcrafted procedure first filters frames according to object visibility and then selects the pseudo coordinate frame based on projected scale and geometric completeness.
However, these proxy criteria do not directly reflect whether a coordinate frame is suitable for downstream 3D grounding. Our reinforcement learning strategy instead evaluates each selected frame according to the grounding IoU achieved under that coordinate condition.
Consequently, the learned selector favors frames containing more useful semantic and geometric cues for localizing the referred object, leading to consistently higher 3D IoU.

\section{Limitations}
\label{sec:limitations}

Despite the strong performance of our framework, two limitations remain.
First, coordinate frame selection is still imperfect.
The large gap between the predicted and oracle frames indicates that the model may select frames with insufficient target visibility or unfavorable spatial configurations, which limits subsequent grounding performance.
Second, the substantial improvement brought by box refinement, especially at $\mathrm{Acc}@0.5$, shows that coordinate-conditioned grounding still struggles with precise 3D box estimation.
Therefore, further improvements require both more reliable coordinate frame selection and stronger fine-grained geometric reasoning for accurate 3D localization.

\bibliography{main}

@String(CVPR  = {IEEE Conf. Comput. Vis. Pattern Recog.})

@String(ICCV  = {Int. Conf. Comput. Vis.})

@String(ECCV  = {Eur. Conf. Comput. Vis.})

@String(NeurIPS = {Adv. Neural Inform. Process. Syst.})

@String(ICML  = {Int. Conf. Mach. Learn.})

@String(ICLR  = {Int. Conf. Learn. Represent.})

@String(AAAI  = {AAAI})

@String(TMLR  = {Trans. Mach. Learn Res.})

@String(CVPR  = {CVPR})

@String(ICCV  = {ICCV})

@String(ECCV  = {ECCV})

@String(NeurIPS = {NeurIPS})

@String(ICML  = {ICML})

@String(ICLR  = {ICLR})

@String(TMLR  = {TMLR})

@inproceedings{
zheng2025learning,
title={{Learning from Videos for 3D World: Enhancing MLLMs with 3D Vision Geometry Priors}},
author={Duo Zheng and Shijia Huang and Yanyang Li and Liwei Wang},
booktitle={Proceedings of the Advances in Neural Information Processing Systems (NeurIPS)},
year={2025},
}

@article{bai2025qwen2,
  title={{Qwen2.5-VL Technical Report}},
  author={Bai, Shuai and Chen, Keqin and Liu, Xuejing and Wang, Jialin and Ge, Wenbin and Song, Sibo and Dang, Kai and Wang, Peng and Wang, Shijie and Tang, Jun and others},
  journal={arXiv preprint arXiv:2502.13923},
  year={2025}
}

@article{bai2025qwen3,
  title={{Qwen3-VL Technical Report}},
  author={Shuai Bai and Yuxuan Cai and Ruizhe Chen and Keqin Chen and Xionghui Chen and Zesen Cheng and Lianghao Deng and Wei Ding and Chang Gao and Chunjiang Ge and Wenbin Ge and Zhifang Guo and Qidong Huang and Jie Huang and Fei Huang and Binyuan Hui and Shutong Jiang and Zhaohai Li and Mingsheng Li and Mei Li and others},
  year={2025},
  journal={arXiv preprint arXiv:2511.21631},
}

@article{shen2025vlm,
  title={{VLM-R1: A stable and generalizable R1-style large vision-language model}},
  author={Shen, Haozhan and Liu, Peng and Li, Jingcheng and Fang, Chunxin and Ma, Yibo and Liao, Jiajia and Shen, Qiaoli and Zhang, Zilun and Zhao, Kangjia and Zhang, Qianqian and others},
  journal={arXiv preprint arXiv:2504.07615},
  year={2025}
}

@article{li2025qinsight,
  title={{Q-Insight: Understanding Image Quality via Visual Reinforcement Learning}},
  author={Li, Weiqi and Zhang, Xuanyu and Zhao, Shijie and Zhang, Yabin and Li, Junlin and Zhang, Li and Zhang, Jian},
  journal={Proceedings of the Advances in Neural Information Processing Systems (NeurIPS)},
  year={2025}
}

@inproceedings{xu2024fakeshield,
    title={{FakeShield: Explainable Image Forgery Detection and Localization via Multi-Modal Large Language Models}},
    author={Xu, Zhipei and Zhang, Xuanyu and Li, Runyi and Tang, Zecheng and Huang, Qing and Zhang, Jian},
    booktitle={International Conference on Learning Representations (ICLR)},
    year={2025}
}

@article{guo2025deepseek,
  title={{DeepSeek-R1 incentivizes reasoning in LLMs through reinforcement learning}},
  author={Guo, Daya and Yang, Dejian and Zhang, Haowei and Song, Junxiao and Wang, Peiyi and Zhu, Qihao and Xu, Runxin and Zhang, Ruoyu and Ma, Shirong and Bi, Xiao and others},
  journal={Nature},
  year={2025},
}

@article{yuan2025scene,
  title={{Scene-R1: Video-Grounded Large Language Models for 3D Scene Reasoning without 3D Annotations}},
  author={Yuan, Zhihao and Jiang, Shuyi and Feng, Chun-Mei and Zhang, Yaolun and Cui, Shuguang and Li, Zhen and Zhao, Na},
  journal={arXiv preprint arXiv:2506.17545},
  year={2025}
}

@article{hurst2024gpt,
  title={{GPT-4o System Card}},
  author={Hurst, Aaron and Lerer, Adam and Goucher, Adam P and Perelman, Adam and Ramesh, Aditya and Clark, Aidan and Ostrow, AJ and Welihinda, Akila and Hayes, Alan and Radford, Alec and others},
  journal={arXiv preprint arXiv:2410.21276},
  year={2024}
}

@article{lillava,
  title={{LLaVA-OneVision: Easy Visual Task Transfer}},
  author={Li, Bo and Zhang, Yuanhan and Guo, Dong and Zhang, Renrui and Li, Feng and Zhang, Hao and Zhang, Kaichen and Zhang, Peiyuan and Li, Yanwei and Liu, Ziwei and Li, Chunyuan},
  journal={Transactions on Machine Learning Research (TMLR)},
  year={2024}
}

@article{LLaVA-OneVision-1.5,
  title={{LLaVA-OneVision-1.5: Fully Open Framework for Democratized Multimodal Training}},
  author={An, Xiang and Xie, Yin and Yang, Kaicheng and Zhang, Wenkang and Zhao, Xiuwei and Cheng, Zheng and Wang, Yirui and Xu, Songcen and Chen, Changrui and Zhu, Didi and others},
  journal={arXiv preprint arXiv:2509.23661},
  year={2025}
}

@inproceedings{chen2020scanrefer,
  title={{ScanRefer: 3D Object Localization in RGB-D Scans Using Natural Language}},
  author={Chen, Dave Zhenyu and Chang, Angel X and Nie{\ss}ner, Matthias},
  booktitle={European Conference on Computer Vision (ECCV)},
  year={2020}
}

@inproceedings{dai2017scannet,
    title={{ScanNet: Richly-annotated 3D Reconstructions of Indoor Scenes}},
    author={Dai, Angela and Chang, Angel X. and Savva, Manolis and Halber, Maciej and Funkhouser, Thomas and Nie{\ss}ner, Matthias},
    booktitle = {Proceedings of the IEEE/CVF Conference on Computer Vision and Pattern Recognition (CVPR)},
    year = {2017}
}

@article{achlioptas2020referit_3d,
    title={{ReferIt3D: Neural Listeners for Fine-Grained 3D Object Identification in Real-World Scenes}},
    author={Achlioptas, Panos and Abdelreheem, Ahmed and Xia, Fei and Elhoseiny, Mohamed and Guibas, Leonidas},
    journal={European Conference on Computer Vision (ECCV)},
    year={2020}
}

@inproceedings{wu2023eda,
  title={{EDA: Explicit Text-Decoupling and Dense Alignment for 3D Visual Grounding}},
  author={Wu, Yanmin and Cheng, Xinhua and Zhang, Renrui and Cheng, Zesen and Zhang, Jian},
  booktitle={Proceedings of the IEEE/CVF Conference on Computer Vision and Pattern Recognition (CVPR)},
  year={2023}
}

@article{hong20233d,
  title={{3D-LLM: Injecting the 3D World into Large Language Models}},
  author={Hong, Yining and Zhen, Haoyu and Chen, Peihao and Zheng, Shuhong and Du, Yilun and Chen, Zhenfang and Gan, Chuang},
  journal={Proceedings of the Advances in Neural Information Processing Systems (NeurIPS)},
  year={2023}
}

@inproceedings{zheng2025video,
  title={{Video-3D LLM: Learning Position-Aware Video Representation for 3D Scene Understanding}},
  author={Zheng, Duo and Huang, Shijia and Wang, Liwei},
  booktitle={Proceedings of the IEEE/CVF Conference on Computer Vision and Pattern Recognition (CVPR)},
  year={2025}
}

@inproceedings{dust3r_cvpr24,
      title={{DUSt3R: Geometric 3D Vision Made Easy}},
      author={Shuzhe Wang and Vincent Leroy and Yohann Cabon and Boris Chidlovskii and Jerome Revaud},
      booktitle = {Proceedings of the IEEE/CVF Conference on Computer Vision and Pattern Recognition (CVPR)},
      year = {2024}
}

@inproceedings{wang2025vggt,
  title={{VGGT: Visual Geometry Grounded Transformer}},
  author={Wang, Jianyuan and Chen, Minghao and Karaev, Nikita and Vedaldi, Andrea and Rupprecht, Christian and Novotny, David},
  booktitle={Proceedings of the IEEE/CVF Conference on Computer Vision and Pattern Recognition (CVPR)},
  year={2025}
}

@inproceedings{wang2024embodiedscan,
  title={{EmbodiedScan: A Holistic Multi-Modal 3D Perception Suite towards Embodied AI}},
  author={Wang, Tai and Mao, Xiaohan and Zhu, Chenming and Xu, Runsen and Lyu, Ruiyuan and Li, Peisen and Chen, Xiao and Zhang, Wenwei and Chen, Kai and Xue, Tianfan and others},
  booktitle={Proceedings of the IEEE/CVF Conference on Computer Vision and Pattern Recognition (CVPR)},
  year={2024}
}

@inproceedings{zhang2023multi3drefer,
  title={{Multi3DRefer: Grounding Text Description to Multiple 3D Objects}},
  author={Zhang, Yiming and Gong, ZeMing and Chang, Angel X},
  booktitle={Proceedings of the IEEE/CVF International Conference on Computer Vision (ICCV)},
  year={2023}
}

@inproceedings{
wang2026pi,
title={{$\pi^3$: Permutation-Equivariant Visual Geometry Learning}},
author={Yifan Wang and Jianjun Zhou and Haoyi Zhu and Wenzheng Chang and Yang Zhou and Zizun Li and Junyi Chen and Jiangmiao Pang and Chunhua Shen and Tong He},
booktitle={International Conference on Learning Representations (ICLR)},
year={2026},
}

@InProceedings{zhu2024llava,
    author    = {Zhu, Chenming and Wang, Tai and Zhang, Wenwei and Pang, Jiangmiao and Liu, Xihui},
    title     = {{LLaVA-3D: A Simple yet Effective Pathway to Empowering LMMs with 3D Capabilities}},
    booktitle = {Proceedings of the IEEE/CVF International Conference on Computer Vision (ICCV)},
    year      = {2025},
}

@inproceedings{zhang2025from,
 author = {Zhang, Jiahui and Chen, Yurui and Xu, Yueming and Huang, Ze and Mei, Jilin and Chen, Chunhui and Zhou, Yanpeng and Yuan, Yu-Jie and Cai, Xinyue and Huang, Guowei and Quan, Xingyue and Xu, Hang and Zhang, Li},
 booktitle = {Proceedings of the Advances in Neural Information Processing Systems (NeurIPS)},
 title = {{From Flatland to Space: Teaching Vision-Language Models to Perceive and Reason in 3D}},
 year = {2025}
}

@inproceedings{
hu2025omniview,
title={{Omni-View: Unlocking How Generation Facilitates Understanding in Unified 3D Model based on Multiview images}},
author={JiaKui Hu and Shanshan Zhao and Qing-Guo Chen and Xuerui Qiu and Jialun Liu and Zhao Xu and Weihua Luo and Kaifu Zhang and Yanye Lu},
booktitle={International Conference on Learning Representations (ICLR)},
year={2026},
}

@article{chen2024grounded3dllm,
      title={{Grounded 3D-LLM with Referent Tokens}},
      author={Chen, Yilun and Yang, Shuai and Huang, Haifeng and Wang, Tai and Lyu, Ruiyuan and Xu, Runsen and Lin, Dahua and Pang, Jiangmiao},
      journal={arXiv preprint arXiv:2405.10370},
      year={2024},
}

@article{huang20253d,
  title={{3D-R1: Enhancing Reasoning in {3D VLMs} for Unified Scene Understanding}},
  author={Huang, Ting and Zhang, Zeyu and Tang, Hao},
  journal={arXiv preprint arXiv:2507.23478},
  year={2025}
}

@article{zhang2025vqinsight,
  title={{VQ-Insight: Teaching VLMs for AI-Generated Video Quality Understanding via Progressive Visual Reinforcement Learning}},
  author={Zhang, Xuanyu and Li, Weiqi and Zhao, Shijie and Li, Junlin and Zhang, Li and Zhang, Jian},
  journal={Proceedings of the AAAI Conference on Artificial Intelligence (AAAI)},
  year={2026}
}

@inproceedings{kingma2014adam,
  title     = {{Adam: A Method for Stochastic Optimization}},
  author    = {Kingma, Diederik P. and Ba, Jimmy},
  booktitle = {International Conference on Learning Representations  (ICLR)},
  year      = {2015},
}

@article{deng2025bagel,
  title   = {{Emerging Properties in Unified Multimodal Pretraining}},
  author  = {Deng, Chaorui and Zhu, Deyao and Li, Kunchang and Gou, Chenhui and Li, Feng and Wang, Zeyu and Zhong, Shu and Yu, Weihao and Nie, Xiaonan and Song, Ziang and Shi, Guang and Fan, Haoqi},
  journal = {arXiv preprint arXiv:2505.14683},
  year    = {2025}
}

@inproceedings{linghu20263d,
  title={{3D-RFT: Reinforcement Fine-Tuning for Video-based 3D Scene Understanding}},
  author={Linghu, Xiongkun and Huang, Jiangyong and Jia, Baoxiong and Huang, Siyuan},
  booktitle={Proceedings of the International Conference on Machine Learning (ICML)},
  year={2026}
}

@inproceedings{kato2023arkitscenerefer,
  title={{ARKitSceneRefer: Text-based localization of small objects in diverse real-world 3D indoor scenes}},
  author={Kato, Shunya and Kurita, Shuhei and Chu, Chenhui and Kurohashi, Sadao},
  booktitle={Findings of the Association for Computational Linguistics: EMNLP 2023},
  year={2023}
}

@inproceedings{
dehghan2021arkitscenes,
title={{ARKitScenes - A Diverse Real-World Dataset for 3D Indoor Scene Understanding Using Mobile RGB-D Data}},
author={Gilad Baruch and Zhuoyuan Chen and Afshin Dehghan and Tal Dimry and Yuri Feigin and Peter Fu and Thomas Gebauer and Brandon Joffe and Daniel Kurz and Arik Schwartz and Elad Shulman},
booktitle={Thirty-fifth Conference on Neural Information Processing Systems Datasets and Benchmarks Track (Round 1)},
year={2021},
}

@article{yu2026dapo,
  title={{DAPO: An Open-Source LLM Reinforcement Learning System at Scale}},
  author={Yu, Qiying and Zhang, Zheng and Zhu, Ruofei and Yuan, Yufeng and Zuo, Xiaochen and Yue, Yu and Dai, Weinan and Fan, Tiantian and Liu, Gaohong and Liu, Lingjun and others},
  journal={Proceedings of the Advances in Neural Information Processing Systems (NeurIPS)},
  year={2025}
}
\end{document}